%% file: main.tex
\documentclass[transmag]{IEEEtran}

\usepackage[utf8]{inputenc}
\usepackage{amsmath}
\usepackage{amssymb}
\usepackage{hyperref}
\usepackage{eqparbox}
\usepackage{latexsym}
\usepackage{graphicx}
\usepackage{float}
\usepackage[dvipsnames]{xcolor}
\usepackage{multirow}
\usepackage{tabularx}
\usepackage{comment}
\usepackage{pgfplots}
\usepackage{pgfplotstable}
\usepackage{booktabs}
\usepackage{xcolor}
\usepackage{listings}
\usepackage{makecell}
\usepackage{pgf}
\usepackage{mdframed}
\usepackage{setspace}
\usepackage{algorithm}
\usepackage{algpseudocode}
\usepackage{mathtools}
\usepackage{array}
\usepackage{circledsteps}
\usepackage[nohyperlinks,printonlyused,withpage]{acronym}
\usepackage{tikz}
\usepackage{longtable}
\usepackage{pdflscape}
\usepackage{pdfpages}
\usepackage{color, colortbl}
\usepackage{amsfonts}
\usepackage{balance}
\usepackage{amsthm}
\usepackage{carshapes}
\usepackage{pbox}
\usepackage{bm}
\usepackage{dblfloatfix}

\def\BibTeX{{\rm B\kern-.05em{\sc i\kern-.025em b}\kern-.08em T\kern-.1667em\lower.7ex\hbox{E}\kern-.125emX}}
\usetikzlibrary{shapes, shapes.geometric, shapes.multipart, shapes.misc, positioning, decorations.pathreplacing, calligraphy, automata, arrows.meta, calc, patterns, trees, backgrounds}

\newcommand{\sunderline}[1]{%
	\underline{\smash{#1}\vphantom{T}}\vphantom{#1}%
}

\definecolor{codegreen}{rgb}{0,0.6,0}
\definecolor{codegray}{rgb}{0.5,0.5,0.5}
\definecolor{codepurple}{rgb}{0.58,0,0.82}
\definecolor{backcolour}{rgb}{0.95,0.95,0.92}
\lstdefinestyle{Py}{
language=python,
basicstyle=\scriptsize\ttfamily,
backgroundcolor=\color{backcolour},   
keywordstyle=\color{codegreen},
numberstyle=\tiny\color{codegray},
stringstyle=\color{codepurple},
breakatwhitespace=false,         
breaklines=true,
captionpos=b,                    
keepspaces=true,
numbers=left,
numbersep=5pt,                  
showspaces=false,                
showstringspaces=false,
showtabs=false,                  
tabsize=2,
morekeywords={self,with,@augment}
}
\theoremstyle{definition}
\newtheorem{definition}{Definition}[section]

\newcommand{\CN}[1]{\textcolor{red}{CN: #1}}
\newcommand{\LW}[1]{\textcolor{violet}{LW: #1}}
\newcommand{\MBu}[1]{\textcolor{green}{MBu: #1}}
\newcommand{\MiS}[1]{\textcolor{orange}{MiS: #1}}

\renewcommand{\CN}[1]{}
\renewcommand{\LW}[1]{}
\renewcommand{\MBu}[1]{}
\renewcommand{\MiS}[1]{}

\acrodef{AUTO}[A.U.T.O.]{Automotive Urban Traffic Ontology}

\begin{document}

\title{Using Ontologies for the Formalization and Recognition of Criticality for Automated Driving%
}

\author{Lukas~Westhofen, %
        Christian~Neurohr, %
        Martin~Butz, %
        Maike~Scholtes, %
        and~Michael~Schuldes%
        \thanks{The research leading to these results is funded by the German Federal Ministry for Economic Affairs and Climate Action within the project 'Verification \& Validation Methods for Automated Vehicles in Urban Environments'. The authors would like to thank the consortium for the successful cooperation.}
        \thanks{Lukas Westhofen and Christian Neurohr are with the German Aerospace Center (DLR) e.V., Institute of Systems Engineering for Future Mobility, Oldenburg, Germany (e-mail: firstname.lastname@dlr.de).}%
        \thanks{Martin Butz is with the Robert Bosch GmbH, Renningen, Germany (e-mail: martin.butz@de.bosch.com).}%
        \thanks{Maike Scholtes is with the RWTH Aachen, Germany (e-mail: maike.scholtes@rwth-aachen.de).}%
        \thanks{Michael Schuldes is with the ika, Aachen, Germany (e-mail: michael.schuldes@ika.rwth-aachen.de).}%
}

\IEEEtitleabstractindextext{
	\begin{abstract}
		Knowledge representation and reasoning has a long history of examining how knowledge can be formalized, interpreted, and semantically analyzed by machines. 
		In the area of automated vehicles, recent advances suggest the ability to formalize and leverage relevant knowledge as a key enabler in handling the inherently open and complex context of the traffic world. 
		This paper demonstrates ontologies to be a powerful tool for a) modeling and formalization of and b) reasoning about factors associated with criticality in the environment of automated vehicles. 
		For this, we leverage the well-known 6-Layer Model to create a formal representation of the environmental context. 
		Within this representation, an ontology models domain knowledge as logical axioms, enabling deduction on the presence of critical factors within traffic scenarios. 
		For executing automated analyses, a joint description logic and rule reasoner is used in combination with an a-priori predicate augmentation.
		We elaborate on the modular approach, present a publicly available implementation, and exemplarily evaluate the method by means of a large-scale drone data set of urban traffic scenarios. 
	\end{abstract}
	
}

\maketitle

\section{INTRODUCTION}
\label{sec:intro}

\input{sections/01_Introduction}

\section{Related Work}
\label{sec:related}

\input{sections/02_Related_Work}

\section{Preliminaries}
\label{sec:approach}

\input{sections/03_Preliminaries}

\section{Reasoning about Criticality for Open Context Systems}
\label{sec:criticality-reasoning}

\input{sections/04_Criticality_Reasoning}

\section{The Automotive Urban Traffic Ontology}
\label{sec:auto}

\input{sections/05_AUTO}

\section{Ontology-Based Formalization of Criticality Phenomena}
\label{sec:formal}

\input{sections/06_Formalization}

\section{Implementing the Criticality Recognition}
\label{sec:eval}

\input{sections/07_Evaluation}

\section{Evaluation and Results}
\label{sec:results}

\input{sections/08_Results}

\section{Conclusion and Future Work}
\label{sec:conc}

\input{sections/09_Conclusion}

\appendix[Evaluated Criticality Phenomena]
\input{sections/10_Appendix}

\bibliographystyle{ieeetr}
\bibliography{IEEEabrv, Literature}

\begin{IEEEbiographynophoto}{Lukas Westhofen} received the degrees of B.Sc. and M.Sc. (with honors) in computer science by the RWTH Aachen University in 2015 and 2019, specializing on the topics of probabilistic programs and software verification. He currently works at the German Aerospace Center (DLR e.V.) Institute of Systems Engineering for Future Mobility, where his research focuses on developing methods to establish confidence in the safety of automated vehicles. More specifically, his interests include the formalization of knowledge as well as its exploitation for safeguarding automated driving functions.
\end{IEEEbiographynophoto}

\begin{IEEEbiographynophoto}{Christian Neurohr} received the B.Sc. and M.Sc. degrees in mathematics in 2011 and 2013 respectively, both from Technische Universit\"at Kaiserslautern, Germany and the Ph.D. degree (Dr. rer. nat.) from Carl von Ossietzky Universit\"at Oldenburg, Germany in 2018. After a short period as a visiting researcher with the MAGMA group at the University of Sydney, he started his occupation as a postdoctoral researcher at the German Aerospace Center (DLR e.V.) Institute of Systems Engineering for Future Mobility where he is working in the area of scenario-based verification and validation of automated vehicles. Specifically, he is the scientific coordinator of the sub-project 'Criticality Analysis' within the project VVMethods.
\end{IEEEbiographynophoto}

\begin{IEEEbiographynophoto}{Martin Butz} received the Ph.D. degree (Dr. rer. nat) in mathematics from the University of Regensburg in 2012. Since then he is working as a research engineer in the field of formal methods and software architecture at corporate research of Robert Bosch GmbH in Renningen, Germany. The main focus of his current research is verification and validation of autonomous driving functions, in particular, the application of formal modeling techniques for open context problems.
\end{IEEEbiographynophoto}

\begin{IEEEbiographynophoto}{Maike Scholtes} received the B.Sc. and M.Sc. degrees in computational engineering science from RWTH Aachen University, Germany, in 2015 and 2017, respectively, and the Ph.D. degree from the Institute for Automotive Engineering (ika), RWTH Aachen University. During semesters abroad she, inter alia, studied at the University of California, San Diego. After working one year in the area of driving simulation at fka SV Inc. in the Silicon Valley, she started her Ph.D. degree. She works on the assessment of automated driving and ADAS with a focus on machine perception. She received the Dean’s List Award for her M.Sc. degree.
\end{IEEEbiographynophoto}

\begin{IEEEbiographynophoto}{Michael Schuldes} received the B.Sc. and M.Sc. degrees in electrical engineering from RWTH Aachen University, Germany, in 2016 and 2019, respectively. He is currently a Research Assistant with the Institute for Automotive Engineering, RWTH Aachen University. His research interests include assessment of autonomous driving functions with a focus on data-driven scenario-based approaches.
\end{IEEEbiographynophoto}

\end{document}

%% file: sections/01_Introduction.tex

\IEEEPARstart{P}{ursuing} the feat of automated driving -- i.e.\ taking human operators out of the control loop -- has sparked scientific interest for several decades. 
Pushed by pioneers like Ernst Dickmanns as early as the 1980s, it peaked in prestigious research projects such as PROMETHEUS \cite{dickmanns_seeing_1994}. 
Despite focused industry efforts and assertive claims, the dream of full driving automation on urban roads has not been realized ever since. 

Currently, systems at SAE Level 3 \cite{sae2021definitions} -- e.g.\ traffic jam assists -- are developed only for highly restricted operational design domains (ODDs). 
This indicates that a key barrier lies within both the complexity and openness of less restricted driving contexts as well as their robust perception and subsequent semantic understanding. 
Consequentially, the development of an automated driving system (ADS) mature enough to overcome these challenges can only be mastered by including stakeholders of a diverse set of disciplines. 

It is hence imperative to identify a common conceptualization of the ODD across all parties involved in the system's development and operation.
Furthermore, this conceptualization must also be comprehensible for machines, as it forms the basis of the interaction for all subsystems of the ADS with the real world. 
While for restricted and highly regulated domains such as highways, a model may be feasibly constructed in a comparatively short time, an urban context has two properties impeding this approach. 
Firstly, it is highly \emph{complex}, meaning that individuals in such scenarios can be of a vast amount of possible types, each of them possessing a multitude of potentially relevant properties possibly with uncountably large value ranges. 
Secondly, the urban context is \emph{open}, meaning that new classes, properties, and behaviors of individuals can emerge rapidly without prior notice or time to respond. 
Therefore, a central question arises: \emph{How to construct a machine-comprehensible model of open and complex driving contexts?}

Earlier work suggests that such operational domains (ODs) can be structured through the identification and subsequent explanation of safety-relevant factors, called criticality phenomena \cite{neurohr2021criticality}. 
This knowledge can then be constituted to a context model, in turn enabling semantic reasoning. 

For example, such reasoning allows for an automated assessment of the situational risk. 
In the area of automated driving, risk evaluations are often performed using criticality metrics \cite{westhofen2021criticality}, e.g. to guide decisions of the driving automation towards states of minimal risk or to derive relevant test cases within a safety case. 
Essentially, such metrics measure 'late' factors that closely precede traffic conflicts, e.g.\ (predicted) small temporal and spatial distances. 
Naturally, the question arises: Can the previously derived criticality phenomena be simultaneously used to approximate the actual risk earlier?

\begin{figure}[htb!]
	\centering
	\input{tikz/cp-metric-chain}
	\caption{
		Exemplary interaction of criticality phenomena and metrics in traffic conflicts, based on an ontology. 
		Criticality phenomena are hatched, metrics are white. TTB is the time to brake, $a_\mathit{req}$ is the maximum required deceleration.
	}
	\label{fig:cp-metric-chain}
\end{figure}
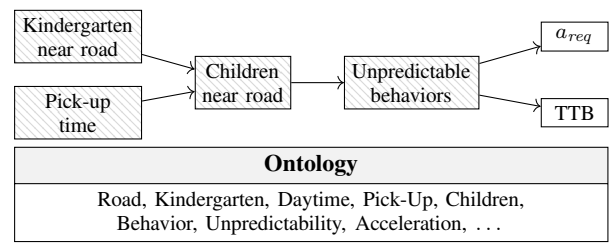

\autoref{fig:cp-metric-chain} sketches an example of how criticality phenomena and metrics interact. 
Here, the presence of kindergartens goes along with the presence of children alongside the driving area at pick-up times. 
The resulting possibility of traffic participants with unpredictable behaviors can be, statistically speaking, antecedent to an increased maximum required deceleration ($a_\mathit{req}$) or decreased time to brake (TTB). 
In total, the recognition and analysis of combinations of factors like the presence of kindergartens, children, and certain daytimes, can lead to a more timely and robust comprehension of the situational risk. 

The work at hand addresses the two challenges directly visible in \autoref{fig:cp-metric-chain}: 
Firstly, building an underlying ontology and secondly, leveraging it to reason on criticality phenomena. 
The core of our work is based on the family of well-understood description logics (DLs) as a formalism to construct an ontology. 
Contemporary approaches suggest the use of ontologies for situation comprehension within ADSs, but lack a justification behind the selection and formalization of relevant concepts. 
As to bridge this gap, we propose to integrate the construction of the ontology into the verification and validation process of an ADS. 
More specifically, this ontology is iteratively refined by expressing the criticality phenomena using DL axioms and rules. 
Finally, the approach enables us to analyze these factors using a statistical evaluation of their association with traffic conflicts as measured by criticality metrics. 

To summarize, the \emph{main contributions} of this work are
\begin{enumerate}
	\item to provide a method that enables the
	\begin{itemize}
		\item construction of an ontology based on a set of criticality phenomena,
		\item formalization of such factors based on the ontology,
		\item inference of their presence in traffic scenarios, and
	\end{itemize}
	\item an open-source implementation of the method, including
	\begin{itemize}
		\item the Automotive Urban Traffic Ontology, and
		\item an ontology-based tooling that recognizes criticality phenomena in traffic scenario data.
	\end{itemize}
\end{enumerate}

After presenting related work in \autoref{sec:related}, \autoref{sec:approach} introduces the foundations of our approach, namely the methodical criticality analysis and the formalisms of description logics and rules. 
The method to both construct an ontology as well as to infer the presence of criticality phenomena (first contribution) is the topic of \autoref{sec:criticality-reasoning}. 
The implemented ontology (second contribution) is portrayed in \autoref{sec:auto}, followed by an example on its iterative construction based on the formalization process in \autoref{sec:formal}. 
To conclude contribution two, \autoref{sec:eval} presents our publicly available tooling. 
An initial evaluation is provided in \autoref{sec:results}, for which we have exemplarily chosen a drone data set, depicting the usage of this method for an in-depth semantic analysis of data.

%% file: tikz/cp-metric-chain.tex
\footnotesize
\begin{tikzpicture}[every node/.style={align=center, draw}]
	\node[text width=1.5cm, pattern=north west lines, pattern color=gray!30]						(kg)	{Kindergarten\\near road};
	\node[below=0.3cm of kg, text width=1.5cm, pattern=north west lines, pattern color=gray!30]		(put)	{Pick-up\\time};
	\node[below right=-0.1 and 0.7cm of kg, pattern=north west lines, pattern color=gray!30]		(cnr) 	{Children\\near road};
	\node[right=0.7cm of cnr, text width=1.6cm, pattern=north west lines, pattern color=gray!30]	(ub)	{Unpredictable\\behaviors};
	\node[right=5.3cm of kg, text width=0.7cm]		(areq)	{$a_\mathit{req}$};
	\node[right=5.3cm of put, text width=0.7cm]		(ttb)	{TTB};
	
	\draw[->] (kg) edge[] (cnr);
	\draw[->] (put) edge[] (cnr);
	\draw[->] (cnr) edge[] (ub);
	\draw[->] (ub) edge[] (areq);
	\draw[->] (ub) edge[] (ttb);
	
	\node[rectangle split, rectangle split parts=2, rectangle split part fill={gray!10, white}, draw, text width=7.7cm, text centered, below=2.1cm of kg.west, anchor=west] (lm) {\small\textbf{Ontology}\nodepart{second}Road, Kindergarten, Daytime, Pick-Up, Children, Behavior, Unpredictability, Acceleration, $\dots$};
\end{tikzpicture}

%% file: sections/02_Related_Work.tex

For road traffic, conceptualizations of knowledge have been collected, aggregated, consolidated, and described since the regulation of the traffic domain. 
For example, in Germany, there exist harmonized taxonomies and guidelines on the type of objects as well as how they shall be constructed \cite{baier2007richtlinien}. 

For automated systems sufficiently intelligent to navigate complex traffic scenarios, the subsequent step is making this aggregated knowledge accessible. 
A unified terminology of the basic models -- scene, situation, scenario -- was defined by Ulbrich et al.\ \cite{Ulbrich2015}. 
Within those scenarios, it was found that classes of entities could be structured along the so-called 6-Layer Model, a deliberately informal but comprehensive framework \cite{scholtes20216}. 
We base the work at hand on these natural-language definitions as well as their classes of entities and provide a formal implementation thereof. 
Going into more detail, Czarnecki et al.\ collected large amounts of structured albeit informal domain knowledge regarding the environment of automated vehicles \cite{Czarnecki2018, Czarnecki2018a}. 
Furthermore, several standards present a structured domain model alongside, e.g.\ ASAM OpenSCENARIO \cite{asamopenscenario}. 
These sources were considered during the development of our implementation. 

Hence, an ontology with formal semantics can be constructed from structured knowledge. 
For automated driving, several groups have formalized their domain conceptualizations by means of ontologies to represent both knowledge and data. 
This includes the \textsc{Ronny} ontology for intersection situations \cite{hummel_description_2009}, the Automotive Global Ontology (AGO) \cite{urbieta2021design}, an Object-Oriented Framework (OOF) \cite{de2022towards}, and an approach of Core Ontologies (CO) \cite{Zhao2015a}. 
Recently, the community initiated standardization efforts for the simulation domain \cite{asamoxo}. 
Three of those ontologies are developed for specific use cases -- intersection recognition for \textsc{Ronny}, closed world scenario descriptions for OOF, and automated decision making for CO -- while the ontology presented for the work at hand aims to support various competencies.  
The AGO provides a knowledge organization framework focused on data labeling, but also includes a methodical approach to derive ontological concepts. 
This derivation is based on pre-existing data schemes to foster compatibility for the subsequent labeling use case. 
Hence, none of the ontologies considers its systematic integration into a safety case by means of a traceable identification and formalization of concepts w.r.t. the safety-relevant aspects of the OD. 
Rather, they represent the particular views and knowledge of the authors and knowledge sources. 

Besides formalizing (shared) conceptualizations, research has leveraged the capabilities using logic reasoning on formalized ontologies to facilitate artificial intelligence functionality in ADSs. 
For a general semantic interpretation of perceived scenes, approaches have been presented as early as 2006 \cite{neumann2006scene}, based on the foundations of prior research in logics for artificial intelligence, including expert systems \cite{russell2002artificial} as well as spatial and temporal logics for robotics \cite{cohn1997qualitative, allen1981interval} in the 1970s to 1990s. 
Applications for automated driving can be traced back to using \textsc{Ronny} on the 2007 DARPA Urban Challenge vehicle AnnieWAY \cite{hummel_description_2009}, which relies on a plain DL approach for scene inference on road geometries. 
We adopt a similar dogma, but also consider scenarios, i.e.\ the dynamical actions of traffic participants, in the inference process. 
Furthermore, our work specifically addresses the problem of lifting concrete domains to abstract ones such that input data becomes suitable for DL reasoning. 
Subsequent to \textsc{Ronny}, various ontology-based approaches for situation comprehension and planning components of automated vehicles were proposed. 
This includes a turning assistant for urban intersections \cite{Boudra2015} and context-aware speed adaption systems \cite{Armand2014, Zhao2015b}. 
Furthermore, ontological formalisms were profoundly leveraged by H\"ulsen et al.\ in order to interpret the driving context at urban intersections using a combined DL-rule approach. \cite{Hulsen2011}. 
Their application of ontology-based inference on A-Boxes is related to the evaluation presented in \autoref{sec:eval}: They specifically leverage the open world assumption and rely on rules and description logics for situation analysis.
Note, however, that the authors were developing their approach specifically for run time inference. 
Reasoning is hence only performed on abstract concepts within a reduced ontology. 
Furthermore, this inference approach is purely scene-based, whereas the work at hand aims to show how inference on large ontologies can be performed on complex scenario data at design time. 

While these use cases relate to the situation comprehension of an ADS, our approach views ontologies as an inherent cornerstone of a safety case, which is in turn required for homologation. 
In the end, a shared ontological model between the stakeholders of the safety case will be necessary. 
For this purpose, it is hence insufficient to omit a justification for the selection of concepts, e.g.\ why to differentiate between a personal mobility device and a bicycle. 
This justification can then be used for verifying and validating the situation comprehension of the ADS. 
First steps have been made by Jatzkowski et al.\, who demonstrated how the relevancy of scene concepts can be derived from abstract system skills \cite{jatzkowski2021knowledge}. 
In order to methodically support a safety case, Bagschik and Menzel examined how a knowledge base can be leveraged for automated scene creation \cite{Bagschik2018a} which in turn can be used to derive test scenarios from keywords \cite{menzel2019functional}. 
In such a scenario-based verification and validation process, it is essential to assign observed concrete scenarios to scenario classes, an approach coined as tagging in prior work \cite{de2020tagging}. 
Albeit all four approaches are methodically integrated into their respective verification and validation processes, they do not highlight the applicability of the underlying logical formalisms for automated reasoning, or leave this issue for future work. 
However, it is only due to a logical foundation that a) concepts become rigorously traceable and b) inference can be leveraged for e.g. formal scenario analysis or consistency checks of the domain model (i.e. whether certain axioms are contradictory). 

Finally, when moving from risk assessment at design time to dynamic risk assessment, early work shows that ontological reasoning can be employed to quantify risk at run time \cite{Mohammad2015}. 
Here, a rudimentary and small ontology was used for demonstration purposes, leaving open the methodical construction of larger ontologies necessary to represent complex urban contexts and the handling of issues accompanied by the growth in size in terminological and assertional components. 


%% file: sections/03_Preliminaries.tex

\subsection{Criticality and its Phenomena within the Open Context}
\label{subsec:criticality}

\paragraph{The Open Context Problem}
An open and complex urban context poses great challenges when aiming to safely deploy an ADS \cite{poddey2019validation, neurohr2021criticality}. 
Therefore, the ISO/DIS 21448 instructs to reduce the set of unknown hazardous scenarios, which can lead to unsafe situations if they are not considered in the design process \cite{iso21448}. 
This requirement becomes harder to fulfill as the context grows in openness and complexity.

\paragraph{Criticality Analysis}
It is therefore imperative to analyze the operational environment prior to system design, which was coined under the term \emph{criticality analysis} \cite{neurohr2021criticality}. 
Its goal is to identify and understand the OD elements that are -- independent of the system's realization -- safety-relevant for the driving task.
Hence, it uncovers both potential unknown unknowns as well as increases the understanding of known unknowns. 
The analysis delivers evidences for OD coverage and decomposition by a) the creation of a finite catalog of abstract scenarios that can be used to identify both requirements on the system behavior and relevant test cases, b) an argumentation about the coverage of such a catalog w.r.t. the OD, and c) a downstream effort reduction by providing generic statements about the safety-relevant elements of the OD.

To this end, one is specifically concerned with identifying situations of high accident risk. 
Hence, criticality can roughly be understood as a function of the probability and severity of the occurrence of \emph{any} accident given a traffic situation. 
We build upon the following definitions from prior work.

\begin{definition}[Criticality]\cite{neurohr2021criticality}
	Criticality is the combined risk of the involved actors when the situation is continued.
\end{definition}

Aspects of criticality are measured using criticality metrics:
\begin{definition}[Criticality Metric]\cite{westhofen2021criticality}
	A criticality metric is a function $\kappa: \mathcal{S} \times \mathbb{R}^+ \rightarrow O$ that measures for a given traffic scene $S \in \mathcal{S}$ at a time $t \in \mathbb{R}^+$ aspects of criticality on a predetermined scale of measurement $O \subseteq \mathbb{R} \cup {-\infty, +\infty}$. Scenario level criticality metrics extend this definition from scenes to scenarios.
\end{definition}

Criticality metrics measure late factors present in the chain of events leading to traffic conflicts. 
Consequentially, a measurement of factors preceding in turn increases in criticality metrics can be fruitful for purposefully generating evidences in a safety case, e.g.\ to identify factors of high relevance. 
We refer to these factors as criticality phenomena.
\begin{definition}[Criticality Phenomenon]\cite{neurohr2021criticality}
	A criticality phenomenon is a single influencing factor, or a combination thereof, that is associated with increased criticality in a scene or scenario.
\end{definition}
This work examines how a formalization of such phenomena can be leveraged for the construction of an ontology as well as for an analysis of criticality in traffic scenarios. 
The construction of such an ontology is inherently tied to a safety case, as elaborated in \autoref{subsec:onto-open-context}. 
A subsequent ontology-based analysis contributes, in turn, to the single steps of a verification and validation process, as illustrated in \autoref{subsec:competency-questions}. 

\subsection{Knowledge Representation and Reasoning}
\label{subsec:krr}

Knowledge representation and reasoning concerns formalisms that model expert knowledge as well as methods thereon that enable machines to reason about conceptualizations of the real world. 
We base our approach on those foundations, which are concisely introduced in the following.

\paragraph{Ontologies and Description Logics}

The term ontology originates from philosophy, denoting the discipline of studying things, their existence, and human conceptualizations thereof. 
Based on this, the fields of formal logic and computer science specify an ontology as a formalization of a conceptualization of a shared mental model of a certain aspect of the real world (often called the \emph{domain of discourse}). 

Such a shared formal model of the domain of discourse enables multiple stakeholder to agree on a common and consistent terminological basis. 
This is specifically important considering the multiple points of failure in information exchange: 
Different symbols can stand for the same real-world things, and different real-world things can be referred to by the same symbols. 
Furthermore, a layer of indirection is introduced by the interpreter through first resolving symbols to certain concepts, which is eventually resolved to a (potentially unintendedly different) real-world object. 
Ontologies make those relations explicit: they clearly assign semantics to concepts which are unambiguously denoted by a set of symbols. 
They can also aid in making the ambiguities of natural language explicit. 
Having a consistent and commonly understood ontology therefore mitigates mismatches in the semantics of used symbols and concepts among all parties. 

DLs is a family of formal logics to represent such ontologies as a decidable fragment of first order logic. 
At the core of DLs are so-called \emph{concepts} (or classes), which are semantically defined as sets of individuals, and \emph{roles} (or relations), which are, analogously, sets of tuples of individuals.
Individuals are concrete, existing instances within the domain of interest \cite{baader2003description}. 
All three can be referred to by their names, i.e.\ symbols. 

\begin{definition}[Vocabulary]
	A vocabulary is a triple $\mathsf{N} = (\mathsf{N}_R, \mathsf{N}_C, \mathsf{N}_I)$ of sets of \emph{role names} $\mathsf{N}_R$, \emph{concept names} $\mathsf{N}_C$, and \emph{individual names} $\mathsf{N}_I$. 
\end{definition}

\begin{definition}[Ontology]
	\label{def:ontology}
	An ontology over a vocabulary $\mathsf{N}$ is a tuple $\mathcal{O} = (\mathcal{T}, \mathcal{A})$, where
	\begin{itemize}
		\item $\mathcal{T}$ is a finite set of terminological assertions (called \emph{T-Box}), namely \emph{general concepts inclusions} (GCIs) such as $\mathtt{C} \sqsubseteq \mathtt{C'}$ for two concept names $\mathtt{C}, \mathtt{C'} \in \mathsf{N}_C$ as well as \emph{role inclusion axioms} (RIAs) such as $\mathtt{r} \sqsubseteq \mathtt{r'}$ for two role names $\mathtt{r}, \mathtt{r'} \in \mathsf{N}_R$, and
		\item $\mathcal{A}$ is a finite set of \emph{concept and role assertions} (CAs and RAs) (called \emph{A-Box}), such as $\mathtt{C}(\mathtt{x})$ and $\mathtt{r}(\mathtt{x}, \mathtt{x'})$ for $\mathtt{x}, \mathtt{x'} \in \mathsf{N}_I$, $\mathtt{C} \in \mathsf{N}_C$, and $\mathtt{r} \in \mathsf{N}_R$. 
	\end{itemize}
\end{definition}

Note that the possible logical operators that can be used to define the right- and left-hand sides of the GCIs and RIAs in $\mathcal{T}$ are determined by the given DL fragment, consisting e.g. of negation ($\neg$), intersection ($\sqcap$), union ($\sqcup$), and universal ($\forall \mathtt{r}.\mathtt{C}$) and existential ($\exists \mathtt{r}.\mathtt{C}$) role quantification. 
Semantically, the operators are defined in terms of an interpretation $\mathcal{I} = (\Delta^\mathcal{I}, \cdot^\mathcal{I})$, where $\Delta^\mathcal{I}$ is the interpretation domain and $\cdot^\mathcal{I}$ an interpretation function assigning each operator a subset of $\Delta^\mathcal{I}$ or $\Delta^\mathcal{I} \times \Delta^\mathcal{I}$. 
For example, we say that $\mathtt{C} \sqsubseteq \mathtt{C'}$ iff for all interpretations $\mathcal{I}$ it holds that $\mathtt{C}^\mathcal{I} \subseteq \mathtt{C'}^\mathcal{I}$. 
$\top, \bot \in \mathsf{N}_C$ are unique names for the top (universally true) and bottom (universally false) concept with $\top^\mathcal{I} = \Delta^\mathcal{I} \wedge \bot^\mathcal{I} = \emptyset\ \forall\ \mathcal{I}$. 
The work omits further details and assumes some familiarity with DLs \cite{baader2003description}. 

For our purposes, we additionally define the special functions $\mathit{con}: \mathcal{T} \rightarrow 2^{\mathsf{N}_C}$, $\mathit{rol}: \mathcal{T} \rightarrow 2^{\mathsf{N}_R}$, and $\mathit{ind}: \mathcal{A} \rightarrow 2^{\mathsf{N}_I}$ which retrieve the set of employed roles, concepts, and individual names from the given GCI, RIA, CA, and RA respectively. 
In conjunction, we say $\mathit{nms} \coloneqq \mathit{rol} \cup \mathit{con} \cup \mathit{ind}$. 
Furthermore, the set of T-Box axioms that involve a concept or a role $\mathtt{X}$ is defined as $\mathit{axs}(\mathtt{X}) \coloneqq \{ \alpha \in \mathcal{T} \,|\, \mathtt{X} \in \mathit{nms}(\alpha) \}$.

\paragraph{Rules}
The modeled axioms of a DL ontology are often enhanced by rules in form of a set of Horn clauses \cite{krotzsch2010description}. 

\begin{definition}[Rule Atom]
	For a given ontology $\mathcal{O}$ over a vocabulary $\mathsf{N}$, a rule atom over a set of variables names $V$ is an expression of the form
	\begin{itemize}
		\item $\mathtt{C}(v)$,
		\item $\mathtt{r}(v, v')$,
		\item $v = v' \circ v''$ or $v \sim v''$ for $\circ \in \{+, -, \cdot, \div\}$ and\\$\sim \in \{<, \leq, =, \neq, \geq, >\}$,
	\end{itemize}
	where $\mathtt{C} \in \mathsf{N}_C$, $\mathtt{r} \in \mathsf{N}_R$, and $v, v' \in V$, $v'' \in V \cup \mathbb{Q}$.
\end{definition}

\begin{definition}[Rule]
	For a given ontology $\mathcal{O}$ over a vocabulary $\mathsf{N}$, a rule over variables $V$ is a Horn clause of a set of antecedent rule atoms $R_A$ over $V$ and precedent rule atoms $R_P$ over $V' \subseteq V$, i.e. of the form $\bigwedge_{r_a \in R_A} r_a \implies \bigwedge_{r_p \in R_P} r_p$. 
	Semantically, they are defined by universal quantification over the variables of $V$. 
\end{definition}


\paragraph{Reasoning}
A reasoner can then deduct new assertions from the given ontology, and is therefore able to check propositions and answer queries. 
For example, it can deduce the axiom $\mathtt{C'}(\mathtt{x})$ from the previously sketched axioms in \autoref{def:ontology}.  
Generally, we denote the entailment of an axiom $\alpha$ from an ontology $\mathcal{O}$ by $\mathcal{O} \models \alpha$. 
Based on this, a subset of relevant decision problems is:
\begin{itemize}
	\item a concept is inconsistent, i.e.\ $\mathtt{C} \sqsubseteq \bot$,
	\item a concept subsumes another, i.e.\ $\mathtt{C} \sqsubseteq \mathtt{C'}$,
	\item two concepts are equivalent, i.e.\ $\mathtt{C} \sqsubseteq \mathtt{C'} \wedge \mathtt{C'} \sqsubseteq \mathtt{C}$ (concisely, $\mathtt{C} \equiv \mathtt{C'}$), and
	\item membership of a given individual to a concept, i.e.\ $\mathtt{C}(\mathtt{x})$.
\end{itemize}
Furthermore, knowledge bases can be queried, i.e.\ retrieving all individuals $\mathtt{x}: \mathtt{C}(\mathtt{x})$. 
These reasoning inquiries are substantiated to our domain of discourse in \autoref{subsec:competency-questions}.

%% file: sections/04_Criticality_Reasoning.tex

\subsection{Ontologies for Structuring an Open and Complex Context}
\label{subsec:onto-open-context}

The necessity of tackling the open context problem has been introduced in \autoref{subsec:criticality}, for which we propose to use ontologies as introduced in \autoref{subsec:krr}. 
In its core, ontologies are based on the so-called \emph{open-world assumption}, stating that if $\mathcal{O} \not\models \alpha$ then $\mathcal{O} \models \neg \alpha$ does not necessarily hold. 
Therefore, ontologies distinguish between the actual truth value of a statement and the information about it that can be inferred from the current knowledge, mitigating the problem of incomplete knowledge about the open context. 
Here, the openness refers to the inherent existence of things possibly unknown at design time, e.g.\ a traffic participant may not be detected by a machine learning perception system due to its anomalistic shape (such as  costumes during carnivals). 
Although a reasoner may find that, based on the knowledge retrieved from the perception system, $\mathcal{O} \not\models (\exists \mathtt{has\_traffic\_entity} . \mathtt{Traffic\_Participant})(\mathtt{s_{i}})$ holds, it will not conclude that no traffic participant is present in the current scene $\mathtt{s_i}$. 
This can be then, for example, be propagated when querying the ontology whether it knows about the absence of vulnerable road users close to the road. 

\begin{figure}[htb!]
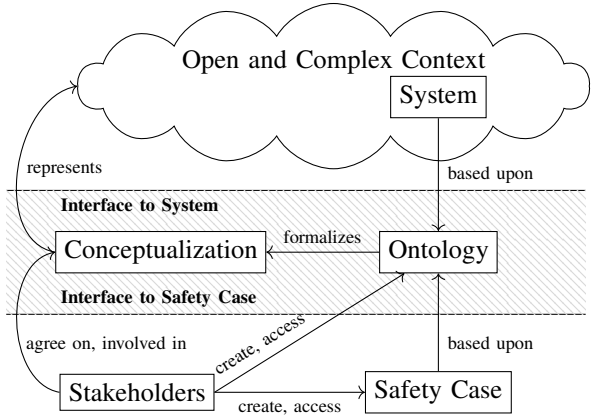

	\centering
	\include{tikz/ontology-open-context}
	\caption{Schematic representation of the role of an ontology in developing and safeguarding complex systems in open contexts. 
	}
	\label{fig:ontology-open-context}
\end{figure}

On a more general note, \autoref{fig:ontology-open-context} schematically depicts how an ontology functions as a central interface between the real world and the various entities that relate to the world. 

In the life cycle of a complex system, those entities are manifold in appearance, ranging from persons, such as developers, engineers, and testers, up to artifacts, such as the safety case, data schemes, and even the system itself, all with diverse models of the world. 
A mismatch in the understanding of the context between one of those entities can be fatal. 
This issue specifically concerns data-focused applications as a well-founded safety case relies not solely on a single data source but rather on various recording modalities to support its claims, e.g. from test vehicles, naturalistic driving studies, or drone data from intersections. 
Their data schemes need to be aligned in order to avoid discrepancies between these evidences. 
For example, a safety case may demonstrate that a machine learning-based pedestrian perception has an acceptable False Negative Rate w.r.t. the overall positive risk balance. 
Everything else should then relate to precisely the same understanding of a pedestrian that was used for labeling the test data and hence for calculating the False Negative Rate. 

Naturally, the questions arise how to reach an agreement on a common conceptualization as well as how to construct a formalization thereof. 
We propose to guide the creation of such a model along the identification of criticality phenomena. 
A methodical criticality analysis iteratively assembles a list of safety-relevant factors within traffic scenarios, although its completeness argumentation is based on two assumptions: 
\begin{enumerate}
	\item there exist finitely many criticality phenomena, and
	\item their traces are recognizable within data.
\end{enumerate}
The justification of these assumptions rests upon a comparison to human driving skills \cite[Sec. IV c]{neurohr2021criticality}. 
The completeness assumption can then be justified as follows: 
By definition, each criticality phenomenon is associated with criticality. 
If we choose a suitable measurement principle (e.g.\ a suitable criticality metric) that can uncover their implications by finding related traffic conflicts, we are able to identify their traces in data. 
By analyzing whether this event can not be explained by a previously discovered phenomenon -- i.e.\ we have found a new phenomenon --, it is possible to achieve a completeness based on the overall finiteness assumption \cite{damm2018exploiting}. 
This completeness can then be recognized by a reduced frequency of unexplained traffic conflicts. 
Further discussion on discovering and explaining phenomena are given by Neurohr et al. \cite{neurohr2021criticality}.

Concurrently to the criticality analysis, a shared ontology can be created from these factors, closing the gap in handling the open context at design time. 
Note that the completeness of the shared ontology is based on the validity of the justification of the two previously mentioned assumptions. 
Obviously, the ontology needs to be maintained during deployment. 
Furthermore, concepts irrelevant for safety, e.g. those regarding mobility or performance goals of the system, may be added. 

\subsection{Competency Questions}
\label{subsec:competency-questions}

We now derive competency questions that aid in developing an ontology for structuring the open context. 

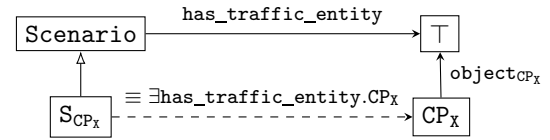
\begin{figure}[htb!]
	\centering
	\input{tikz/high-level-dl}
	\caption{Model of the relevant concepts and relations of the competency questions. A solid edge with solid tip is a custom relation, a solid edge with open tip is the subsumption relation, and a dashed edge is a definition.}
	\label{fig:high-level-dl}
\end{figure}

\paragraph{Preliminaries for Competencies}
First, we introduce some high-level DL notations on the representation of criticality phenomena and scenarios. 
Those high-level classes and their relations are depicted in \autoref{fig:high-level-dl}. 

We start with the concept of all scenarios, $\mathtt{Scenario}$. 
Scenarios can have various member individuals $\mathtt{i}$, denoted by the relation $\mathtt{has\_traffic\_entity}$. 
Criticality phenomena are modeled as sets of such scenario entities.
For each criticality phenomenon $X$, we denote its corresponding concept $\mathtt{CP_X} \sqsubseteq \exists \mathtt{has\_traffic\_entity}^{-1}.\top$. 
For example, the class of all fast vehicles can be defined as $\mathtt{CP_{fast}} \equiv \mathtt{Vehicle} \sqcap \mathtt{Fast\_Object}$. 
Whereas this example solely refers to one subject w.r.t. criticality, such phenomena can generally address one or more objects in addition to the subject. 
Occlusion, for example, has the occluded entity as its subject but also relates to the occluding entity as well as the entity for which the occlusion happens. 
Therefore, objects are referred to by the relation $\mathtt{object_{CP_X}}$, mapping from $\mathtt{CP_X}$ to a number of objects. 

Eventually, we are interested in decomposing the scenario space into a set of abstract scenarios covering the identified criticality phenomena. 
For a given $\mathtt{CP_X}$, we therefore denote its scenario class by $\mathtt{S_{CP_X}} \equiv \exists \mathtt{has\_traffic\_entity}. \mathtt{CP_X}$.

\paragraph{Derivation of Competency Questions}
Starting early in the design phase of ADSs, contemporary approaches suggest to cluster the scenario space into abstract scenario classes. 
This clustering can be formally analyzed to support the experts decomposing the OD. 
Relevant questions are:
\begin{itemize}
	\item Whether a set of scenario classes can occur at the same time -- if they are disjoint, e.g. risk estimations can be performed without explicitly considering their conjunction -- i.e.\ $\mathtt{S_{CP_1}} \sqcap \dots \sqcap \mathtt{S_{CP_n}} \sqsubseteq \bot$ (T1).
	\item Whether a scenario class is subsumed by another, as to facilitate a hierarchical decomposition within the safety argumentation, i.e.\ $\mathtt{S_{CP_1}} \sqsubseteq \mathtt{S_{CP_2}}$ (T2).
	\item Whether the chosen decomposition is complete, i.e.\ $\mathtt{Scenario} \sqsubseteq \mathtt{S_{CP_1}} \sqcup \dots \sqcup \mathtt{S_{CP_n}}$ (T3).
\end{itemize}

This decomposition can be supported by a counterexample-guided abstraction refinement as proposed by Neurohr et al. \cite[Sec. V A 5)]{neurohr2021criticality}. 
For this, is is necessary to assess whether a given scenario is member of a certain scenario class or an individual is or is involved in a criticality phenomenon -- i.e., $\neg \mathtt{S_{CP_X}}(\mathtt{s})$ (A1), $\neg \mathtt{CP_X}(\mathtt{i})$ (A2), or $\exists\mathtt{object_{CP_X}}^{-1}.\top(\mathtt{i})$ (A3). 
Reasoning services on the ontology can support answering such questions formally by delivering counterexamples as to why these memberships do not hold.
Furthermore, data recording and subsequent scenario mining is often used to identify relevant test scenarios. 
Given a real world traffic recording, we can find representative instances for our previously defined abstract scenario classes by asking a reasoner to query for A1 to A3. 
Finally, during system testing, answering A1 to A3 allows to check whether a concrete simulation or test run is an instance of a given test case. 
As one central element of a safety case, system testing can contribute valuable evidences for a necessary homologation prior to deployment. 

However, even after successful homologation, constant monitoring of the ODD is required post-deployment, e.g.\ to detect changes in the assumptions on the ODD that were made in the pre-deployment safety case. 
The ontology can be used to formalize such assumptions on the ODD definition. 
An ontology-based monitor can then decide whether a) the observed scenario is consistent with the modeled knowledge about the world (i.e.\ the assumptions) and b) the observed or predicted future scenario still lies within the ODD. 
Otherwise, appropriate mechanisms have to be executed by the ADS.
The first query reduces to the formal expression $\mathcal{O} \models \mathtt{Scenario}(\mathtt{s})$ (A4), whereas the second one can be based on A1 to A3. 
Especially after incidents but also in joint operation with humans, e.g.\ for Level 3 functions, the system's behavior needs to be explainable at runtime. 
If a vehicle chooses a behavior based on specific circumstances in its environment (i.e.\ assesses A1 to A3), it is advisable to investigate why this behavior was chosen. 
We can query a reasoner to derive a logical explanation as to why $\mathtt{S_{CP_X}}$ was deemed to be present. 

\begin{table*}[htb!]
	\caption{Competency questions of the proposed ontology. Questions with bold-faced identifiers are focused on in this work.}
	\label{tab:competency_questions}
	{
		
		\centering
		\small
		\renewcommand{\arraystretch}{1.5}
		\begin{tabularx}{\textwidth}{p{0.07\textwidth}p{0.21\textwidth}p{0.28\textwidth}p{0.30\textwidth}p{0.03\textwidth}}
			\toprule
			\textbf{Reasoning} & \textbf{DL notation} & \textbf{Explanation} & \textbf{Example} & \textbf{ID} \\
			\midrule
			& $\mathtt{S_{CP_X}}(\mathtt{s})$ & Membership of a scenario $\mathtt{s}$ within a scenario class $\mathtt{S_{CP_X}}$ & Does the scenario have an occluded pedestrian? & \textbf{A1} \\ 
			\multirow{4}{*}{\emph{A-Box}} & $\mathtt{CP_X}(\mathtt{i})$ & Membership of an individual $\mathtt{i}$ within a criticality phenomenon $\mathtt{CP_X}$ & Is the pedestrian occluded? & \textbf{A2} \\ 
			& $\exists\mathtt{object_{CP_X}}^{-1}.\top(\mathtt{i})$ & Membership of an individual $\mathtt{i}$ as an object of a criticality phenomenon $\mathtt{CP_X}$ & In the given scenario, is the ego vehicle causing an occlusion? & \textbf{A3} \\ 
			& $\mathcal{O} \models \mathtt{Scenario}(\mathtt{s})$ & Consistency of an observed scenario $\mathtt{s}$ with the modeled knowledge & Is the perceived pedestrian with 40km/h consistent with the modeled knowledge? & A4 \\
			\midrule
			& $\mathtt{S_{CP_1}} \sqcap \dots \sqcap \mathtt{S_{CP_n}} \sqsubseteq \bot$ & Compatibility of a set of criticality phenomena & Can the criticality phenomena Foggy Night \& Sun Glare exist simultaneously? & T1 \\ 
			\emph{T-Box} & $\mathtt{S_{CP_1}} \sqsubseteq \mathtt{S_{CP_2}}$ & Abstraction relation of criticality phenomena & Is the concretization of Occluded Road User into Occluded Vehicle and Occluded Pedestrian correct? & T2 \\
			& $\mathtt{Scenario} \sqsubseteq \mathtt{S_{CP_1}} \sqcup \dots \sqcup \mathtt{S_{CP_n}}$ & Completeness of a set of scenario classes & Is the scenario space decomposition into right and left turn maneuvers complete? & T3 \\
			\bottomrule
		\end{tabularx}
	}
\end{table*}

\autoref{tab:competency_questions} presents the derived seven competency questions -- A1 to A4 and T1 to T3 -- the proposed ontology was implemented against. 
Note that there also exist other possible competency questions, such as model finding -- algorithmically generating a witness for the satisfiability of $\top \sqsubseteq \neg \texttt{C}$, e.g.\ a concrete scenario $\mathtt{s}$ for a given abstract scenario class $\mathtt{S_{CP_X}}$. 
Such competency questions require different architectures and approaches in ontology design and were thus excluded. 

Whereas the ontology has been implemented w.r.t. all competency questions, the evaluation of the ontology-based tooling as presented in \autoref{sec:eval} and \autoref{sec:results} is focused on the first three competency questions, A1 to A3. 
This allows, on one hand, to keep the ontology implementation generic and applicable to a variety of use cases, and, on the other hand, the practical evaluation focused and comprehensible. 

\subsection{Methodical Formalization and Recognition Criticality}
\label{subsec:general-method}

We have introduced the working hypothesis that by creating an ontology via a set of identified criticality phenomena we are able to achieve an iterative completeness of the model w.r.t. the safety relevant factors. 
Furthermore, we established the practical relevance of the guiding competency questions regarding the safety assurance of ADSs within a thereby created model. 
In particular, we demonstrated the importance of answering competency question A1 -- scenario membership.
Based on these foundations, we propose a general method that
\begin{enumerate}
	\item formalizes criticality phenomena within an iteratively refined ontology, 
	\item transforms data from various sources into the language of the ontology, and
	\item utilizes DL \& rule reasoners to infer the presence of criticality phenomena in such traffic scenarios
\end{enumerate}
 
\begin{figure}[htb!]
	\scalebox{1.0}{\input{tikz/formalization_workflow}}
	\caption{The general method of formalizing a natural language phenomenon and reasoning about its presence in data.}
	\label{fig:general_process}
\end{figure}
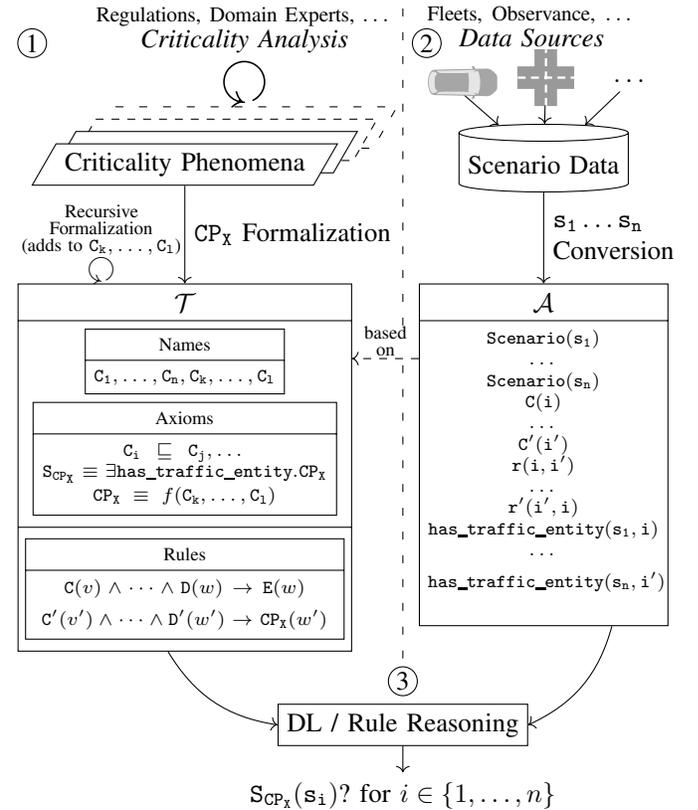

The process is depicted in \autoref{fig:general_process} where these steps are respectively marked by \Circled{1}, \Circled{2}, and \Circled{3}.

In step \Circled{1}, the T-Box $\mathcal{T}$ is iteratively constructed by formalizing a list of criticality phenomena. 
This list is constantly growing by continuous conduction of a criticality analysis \cite{neurohr2021criticality}. 
We therefore pick a criticality phenomenon $\mathtt{CP_X}$ and create a formalization of its semantics using appropriate rules or DL axioms. 
This semantics can be based on both existing concepts -- $\mathtt{C_1}, \dots, \mathtt{C_n}$ -- as well as newly introduced concepts -- $\mathtt{C_k}, \dots, \mathtt{C_l}$. 
This necessitates the addition of $\mathtt{C_k}, \dots, \mathtt{C_l}$ to $\mathcal{T}$, in turn recursively formalizing them until the valid assumption can be made that so-called basic concepts can not be further specified. 
A validation of the assumption is required, i.e.\ it has to be argued why neither a rule nor a description logic axiom is suitable to represent the given concept. 

If one aims to infer the presence of a formalized criticality phenomenon, these basic concepts are then required to be either directly quantifiable or computationally derivable from the data. 
In that case, they shall have a definition in natural language as to facilitate a correct observation. 
If basic concepts are not observable in data (for example, if the recording modality does not allow to recognize certain features such as whether turning indicators are enabled), an analysis of the phenomena relying on those concepts can not be performed. 

Hinging on the assumption of convergence of the criticality analysis, this process leads to an iterative completion of the ontology towards an exhaustive representation of all concepts that are deemed essential for identifying safety-relevant factors.  
This step is elaborated on in detail in the upcoming \autoref{sec:auto} -- concerning the basic ontological framework -- and \autoref{sec:formal} -- concerning the formalization of criticality phenomena within this ontological framework. 

Step \Circled{2} unifies the various scenario data sources by conversion into a $\mathcal{T}$-compliant A-Box $\mathcal{A}$. 
This creates scenarios $\mathtt{s_1}$ to $\mathtt{s_n}$ with traffic entities $\mathtt{i}$ to $\mathtt{i'}$ that are represented using the concepts $\mathtt{C}$, $\mathtt{C'}, \dots$ and roles $r$, $r'$, $\dots$ of $\mathcal{T}$. 
A realization of this step is given in \autoref{subsec:omega} and \autoref{subsec:onto-map}.

Once the manifold data sources are convertible from their native data schemes, new facts can be inferred by performing reasoning on the ontology $\mathcal{O} = (\mathcal{T}, \mathcal{A})$, as indicated by \Circled{3}. 
Such a reasoning pipeline has been prototypically implemented for the work at hand whilst taking  into account various practical concerns, e.g.\ temporal reasoning performance and concrete domain abstraction. 
This finally answers competency question A1, i.e.\ for which $i \in \{1, \dots, n\}$ $\mathtt{S_{CP_X}}(\mathtt{s_i})$ holds, and, per definition, also A2 to A3, cf. \autoref{subsec:ind-cps}. 

%% file: tikz/ontology-open-context.tex
\begin{tikzpicture}
	
	\node[draw=black, align=center, cloud, cloud puffs=14,cloud puff arc=120, aspect=5, inner sep=0] 	(context)	 	{Open and Complex Context\\\vspace{0.3cm}};
	\node[draw=black, align=center, below right=-0.8cm and -1.3cm of context]		(system)	 	{System};
	
	
	\node[below left=0.6cm and 2.3cm of context] (di11) {};
	\node[below right=0.6cm and 1.3cm of context] (di12) {};
	\draw[-] (di11) to (di12);
	
	
	\node[below=1.4cm of di11] (di21) {};
	\node[below=1.4cm of di12] (di22) {};
	\draw[-] (di21) to (di22);
	\path[pattern=north west lines,pattern color=gray!30] (di11.east) to [] (di12.west) to [] (di22.west) to [] (di21.east) to [] (di11.east);
	
	
	\node[below right=-0.15cm and 0.6 of di11]		(i1)	 		{\scriptsize \textbf{Interface to System}};
	\node[above right=-0.15cm and 0.6 of di21]		(i2)	 		{\scriptsize \textbf{Interface to Safety Case}};
	
	
	\node[draw=black, align=center, below=1.5cm of system,fill=white] 			(ontology)					{Ontology};
	\node[draw=black, align=center, left=1.5cm of ontology,fill=white] 			(conceptualization)			{Conceptualization};
	
	
	\node[draw=black, align=center, below=1.3cm of ontology]	(safetycase)		{Safety Case};
	\node[draw=black, align=center, left=2cm of safetycase]	(stakeholders)		{Stakeholders};


	\path[->] (system) edge node[right] {\scriptsize based upon} (ontology);
	\path[->] (safetycase) edge node[right, yshift=-0.3cm] {\scriptsize based upon} (ontology);
	\path[->,out=170, in=190] (stakeholders.west) edge node[right,yshift=-0.3cm] {\scriptsize agree on, involved in} (conceptualization.west);
	\path[->] (ontology) edge node[above] {\scriptsize formalizes}   (conceptualization);
	\path[->,out=170, in=190] (conceptualization.west) edge node[right] {\scriptsize represents}   (context.west);
	\path[->] (stakeholders) edge node[below] {\scriptsize create, access} (safetycase);
	\path[->] (stakeholders.east) edge node[sloped,above,xshift=-0.64cm] {\scriptsize create, access} (ontology);
	
\end{tikzpicture}

%% file: tikz/high-level-dl.tex
\begin{tikzpicture}
	\node (sc)		[rectangle,draw]						{$\mathtt{Scenario}$};
	\node (top)		[rectangle,draw,right=3.65cm of sc]		{$\top$};
	\node (sccp)	[rectangle,draw, below=0.6cm of sc]		{$\mathtt{S_{CP_X}}$};
	\node (cp)		[rectangle,draw, right=4cm of sccp]		{$\mathtt{CP_X}$};
	
	\path [-stealth] (sc) edge node[sloped,above] {\footnotesize$\mathtt{has\_traffic\_entity}$}   (top);
	\path [-stealth] (cp) edge node[right] {\footnotesize$\mathtt{object_{CP_X}}$}   (top);
	\path [-{Latex[open]}] (sccp) edge node[sloped,above] {}   (sc);
	\path [-stealth,dashed] (sccp) edge node[sloped,above] {\footnotesize$\equiv\exists \mathtt{has\_traffic\_entity}. \mathtt{CP_X}$}   (cp);
\end{tikzpicture}

%% file: tikz/formalization_workflow.tex
\begin{tikzpicture}[]
	
	\node[draw=black, align=center, loosely dashed, trapezium, trapezium left angle=60, trapezium right angle=120] 	(cps0)	 		{\phantom{Criticality Phenomena}};
	\node[draw=black, align=center, fill=white, dashed, trapezium, trapezium left angle=60, trapezium right angle=120, below left=-0.4cm and -0.3cm of cps0] 	(cps1)	 		{\phantom{Criticality Phenomena}};
	\node[draw=black, align=center, fill=white, trapezium, trapezium left angle=60, trapezium right angle=120, below left=-0.4cm and -0.3cm of cps1] 	(cps2)	 		{\phantom{Criticality Phenomena}};
	\node[draw=black, align=center, fill=white, trapezium, trapezium left angle=60, trapezium right angle=120, below left=-0.4cm and -0.3cm of cps2] 	(cps)	 		{Criticality Phenomena};
	\node[inner sep=0.3mm, above left=0.7cm and 2.4cm of cps0, draw=black, circle, fill=white] (1) {1};

	\node[above=-0.08cm of cps0, align=center] (critana) {{\footnotesize Regulations, Domain Experts, $\dots$}\\[-0.08cm]\emph{Criticality Analysis}\\[-0.17cm]\rotatebox{180}{\Huge$\circlearrowleft$}};
	
	\node[rectangle split, rectangle split parts=3, draw, text width=4.2cm, text centered, below=1.3cm of cps] (tbox) {$\mathcal{T}$\nodepart{second}\vspace{2.5cm} \nodepart{third} \vspace{1.4cm}};
	
	\node[above left=-0.1cm and -2.3cm of tbox, align=center] (rec) {\scriptsize \color{black}Recursive\\[-0.2cm]\scriptsize Formalization\\[-0.2cm]\scriptsize \color{black}(adds to $\mathtt{C_k}, \dots, \mathtt{C_l}$)\\[-0.3cm]\rotatebox{180}{\Large\color{black}$\circlearrowleft$}};
	
	\node[rectangle split, rectangle split parts=2, draw, text centered, below=0.6cm of tbox.north] (concepts) {\scriptsize Names \nodepart{second}\scriptsize $\mathtt{C_1}, \dots, \mathtt{C_n}, \mathtt{C_k}, \dots, \mathtt{C_l}$};
	
	\node[rectangle split, rectangle split parts=2, draw, text centered, text width=3.8cm, below=0.1cm of concepts] (axioms) {\scriptsize Axioms \nodepart{second}\scriptsize $\mathtt{C_i} \sqsubseteq \mathtt{C_j}, \dots$\\\color{black}$\mathtt{S_{CP_X}} \equiv \exists \mathtt{has\_traffic\_entity}. \mathtt{CP_X}$\\[-0.1cm]\color{black}$\mathtt{CP_X} \equiv f(\mathtt{C_k}, \dots, \mathtt{C_l})$};
		
	\node[rectangle split, rectangle split parts=2, draw, text centered, text width=4cm, below = 0.3 cm of axioms] (rules) {\scriptsize Rules \nodepart{second}\scriptsize $\mathtt{C}(v) \wedge \dots \wedge \mathtt{D}(w) \rightarrow \mathtt{E}(w)$\\\color{black}$\mathtt{C'}(v') \wedge \dots \wedge \mathtt{D'}(w') \rightarrow \mathtt{CP_X}(w')$};
	
	\path[->,black] (cps) edge node[right] {\color{black}$\mathtt{CP_X}$ Formalization} (tbox);
	
	
	\node[above right=1.3cm and 1.7cm of cps0] (div1) {};
	\node[below=8.7cm of div1] (div2) {};
	\draw[loosely dashed] (div1) -- (div2);
	
	\node[right=1.76cm of cps, cylinder, draw=black, shape border rotate=90,shape aspect=0.1, minimum width=1.5cm, minimum height=0.9cm] (data) {Scenario Data};
	\path[->] (2.9,0.45) edge node[] {} (data);
	\node[sedan top,body color=gray!70,window color=gray!20, minimum width=1cm] at (2.9,0.6) {};
	\node[cross out, rotate=45, draw=gray!80, minimum size=5pt, above right=0.25cm and -0.5cm of data, line width=0.3cm, minimum width=0.2cm, minimum height=0.2cm] (crossing) {};
	\node[cross out, rotate=45, draw=white, minimum size=5pt, below right=0cm and -0.75cm of crossing, line width=0.03cm, minimum width=0.5cm, minimum height=0.5cm, densely dashed] (crossingwhite) {};
	\node[above=0.1cm of crossing, text width=5.6cm, align=center] (datasource) {{\footnotesize Fleets, Observance, $\dots$}\\[-0.08cm]\emph{Data Sources}};
	
	\path[->] (crossing) edge node[] {} (data);
	\node[below right=-0.1cm and 0.4cm of crossing] (dots) {$\dots$};
	\path[->] (dots) edge node[] {} (data);
	
	\node[inner sep=0.3mm, right=4.85cm of 1, draw=black, circle, fill=white] (2) {2};
	
	\node[rectangle split, rectangle split parts=2, draw, text width=3.07cm,text centered, anchor=north] (abox) at (tbox.north -| data) {$\mathcal{A}$ \nodepart{second}\scriptsize $\mathtt{Scenario}(\mathtt{s_1})$\\$\dots$\\$\mathtt{Scenario}(\mathtt{s_n})$\\$\mathtt{C}(\mathtt{i})$\\$\dots$\\$\mathtt{C'}(\mathtt{i'})$\\$\mathtt{r}(\mathtt{i}, \mathtt{i'})$\\$\dots$\\$\mathtt{r'}(\mathtt{i'},\mathtt{i})$\\$\mathtt{has\_traffic\_entity}(\mathtt{s_1}, \mathtt{i})$\\$\dots$\\$\mathtt{has\_traffic\_entity}(\mathtt{s_n}, \mathtt{i'})$\vspace{0.33cm}};
	\path[->] (data) edge node[right,align=left] {$\mathtt{s_1} \dots \mathtt{s_n}$\\Conversion} (abox);
	
	\path[->,dashed] ([yshift=-1cm]abox.north west) edge node[align=center,above,fill=white, inner sep=0,yshift=0.03cm] {\scriptsize based\\[-0.2cm]\scriptsize on} ([yshift=-1cm]tbox.north east);
	
	
	\node[rectangle, draw, below=0.3cm of div2] (reasoner) {DL / Rule Reasoning};
	\node[inner sep=0.3mm, above=0.1cm of reasoner, draw=black, circle, fill=white] (3) {3};
	\path[->] (tbox) edge[bend right=25] (reasoner.west);
	\path[->] (abox) edge[bend left=25] (reasoner.east);
	\node[below=0.35cm of reasoner] (result) {${\color{black}\mathtt{S_{CP_X}}}(\mathtt{s_i})$? for $i \in \{1, \dots, n\}$};
	\path[->] (reasoner) edge (result);

\end{tikzpicture}

%% file: sections/05_AUTO.tex

Starting with step \Circled{1} of the sketched method of \autoref{fig:general_process}, this section presents the employed T-Box. 
It describes the conceptual foundations, general structure, idea, modeling, and implementation of an integrated set of suitable ontologies. 

\subsection{The 6-Layer Model as a Conceptual Foundation}

As a tool for structured environment description, the well-known Layer Model can be utilized to create a formal model of the context. 
It was originally defined by Schuldt \cite{Schuldt2017} and has been continuously refined for specific use cases, e.g.\ by Bagschick et al.\ for the instantiation of highway scenes \cite{Bagschik2018a}. 
In a recent work by Scholtes et al., their ideas are examined in-depth and adapted for urban environments while keeping compatibility to the fundamental ideas behind the Layer Model \cite{scholtes20216}. 
Since Scholtes et al. give detailed definitions of the layers as well as justifications, guidance, and examples, it serves as a suitable basis for our use case -- namely, the construction of a formal ontology implementing the Layer Model. 

The work categorizes the environment, its entities, and their properties in six individual layers: Road Network and Traffic Guidance Objects (L1), Roadside Structures (L2), Temporary Modifications of the former (L3), Dynamic Objects (L4), Environmental Conditions (L5), and Digital Information (L6), encompassing the so-called 6-Layer Model (6LM). 
It thereby provides a generally usable, unbiased, and objective environment description, i.e. it functions as a clearly structured basis for scenario descriptions and their ontologies. 
While the 6LM provides a unique top-level structuring for the environment, an ontology details and formalizes the different entities, properties, and relations in a more comprehensive way.

\subsection{Architecture}

In order to address the issue of integrating the 6LM into a formal ontology as necessitated by the process of \autoref{fig:general_process}, we present the \acl{AUTO} (\acs{AUTO}).
In its core, it is a set of integrated ontologies with the goal to enable reasoning about criticality in urban traffic scenarios. 
Due to this scope, the 6LM as defined by Scholtes et al.\ for urban environments, is used as an architectural guiding pattern. 


Architecturally, we separate each layer into a \emph{core} and an \emph{implementation}. 
Cores represent an ontological interface which other layers and applications of the ontology can access without being concerned with detailed domain knowledge, which in turn is given in a set of implementations. 
For example, the concept $\mathtt{Lane}$ is part of the Layer 1 core, but concrete specializations of lanes can be given in an implementation. 

Moreover, this work extends the 6LM by detaching conceptually separated parts which semantically span over multiple layers or are completely independent of the traffic domain.
This modularity enables separation of concerns and re-use of concepts, e.g. for spatial properties (heights, distances, etc.). 

Note that often, so-called Upper Level Ontologies are employed for the purpose of unification. 
We deliberately omit the use of such ontologies as we were not able to identify an ontology that satisfied the needs imposed by the competency questions. 
Namely, this concerns issues of suitable formalization to enable inferences by DL reasoners, documentation, implementation, comprehensibility, and modularity. 
This problem is currently addressed in the standardization project ASAM OpenXOntology \cite{asamoxo}, where an Upper Level Ontology is laboriously integrated into a traffic domain ontology. 

We hence opt for a lean approach, where single, re-usable domain ontologies are imported, as displayed in \autoref{tab:related-ontologies}. 

\begin{table}[htb!]
	\caption{Detached, related ontologies of the 6LM used within \acs{AUTO}}
	\label{tab:related-ontologies}
	\scriptsize
	\centering
	\setlength{\tabcolsep}{2pt}
	\begin{tabular}{llll}
		\toprule
		\textbf{Ontology} & \textbf{Competency} & \textbf{Example Concepts} & \textbf{Example Roles} \\
		\midrule
		\textbf{Time} & Temporal concepts & Intervals, Time Points & After, Before \\
		\textbf{Geometry} & Geometries \& shapes & Geometry, Rectangle & Contain, Disjoint \\
		\textbf{Physics} & Physical properties & Dynamical object & Speed, Acceleration  \\
		\textbf{Act} & Activities over time & Actor, Event, Activity & Conduct, Participate\\
		\textbf{Communication} & Message exchange & Sender, Receiver & Transmit, Receive \\
		\textbf{Perception} & Perceiving Entities & Observer, Occlusion & Is Occluded For \\
		\bottomrule
	\end{tabular}
\end{table}

\begin{figure}[htb!]
	\centering
	\input{tikz/auto-architecture}
	\caption{The modular architecture of the \acl{AUTO}.}
	\label{fig:auto}
\end{figure}
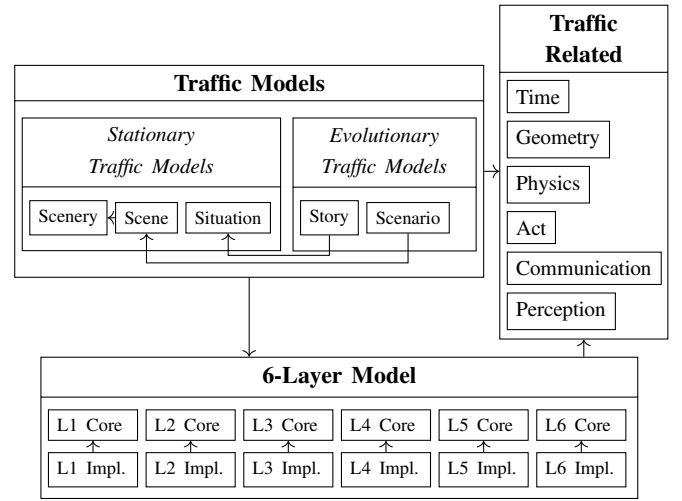

On top of the related ontologies, we introduce a framework in which traffic entities of the 6LM are integrated into traffic models such as scenes and scenarios. 
Collectively, this conglomerate of ontologies yields the \acs{AUTO} shown in \autoref{fig:auto}. 

The diagram indicates the direction of the dependencies of the modules through arrows. 
Therefore, both the traffic models as well as the 6LM rely on the set of related ontologies. 
The former uses e.g.\ temporal concepts to specify scenarios, the latter uses e.g.\ physical properties for traffic participants. 

\subsection{Implementation}

We provide an open-source implementation\footnote{\url{http://github.com/lu-w/auto}} of the previously introduced ontology in OWL2 (i.e. the DL  $\mathcal{SROIQ}^{(D)}$). 

The provided \acs{AUTO} consists of 26 OWL2 files implementing the modular architecture of \autoref{fig:auto}. 
Each file represents one ontology module in which its respective domain concepts are defined. 
For example, the Physics ontology defines concepts such as $\mathtt{Spatial\_Object}$, $\mathtt{Dynamical\_Object}$, and $\mathtt{Non\_Moving\_Dynamical\_Object}$ alongside their natural language definition. 
Here, a $\mathtt{Spatial\_Object}$ is defined as 'any physical object with a geometrical representation, e.g.\ a vehicle'. 
If expressible in DLs, the OWL2 files also specifies logical axioms implementing these natural language definitions. 
For example, the Physics ontology states $\mathtt{Dynamical\_Object} \sqsubseteq \mathtt{Spatial\_Object}$ and $\mathtt{Non\_Moving\_Dynamical\_Object} \equiv \mathtt{Dynamical\_Object} \sqcap \exists \mathtt{has\_velocity}.\{0.0\}$. 
Those logical T-Box axioms then allow the reasoner, in turn, to infer entailed statements on the A-Boxes for the subsequent evaluation -- such as, whether a given vehicle is a $\mathtt{Non\_Moving\_Dynamical\_Object}$. 
For a visual excerpt of the T-Box of \acs{AUTO}, including the subsumption relation and other logical axioms, we reference to \autoref{fig:cp-ex-subsumption}.

We also rely on well-established ontologies, enabling compatibility to a wide range of tooling such as GraphDB:
\begin{itemize}
	\item geometrical concepts are represented using OGC's \emph{GeoSPARQL} ontology\footnote{\url{https://www.ogc.org/standards/geosparql}}, which gives access to DE-9IM and RCC8 vocabularies \cite{hogenboom2010spatial}, and
	\item temporal concepts are represented using W3C's \emph{OWL Time} ontology\footnote{\url{https://www.w3.org/TR/owl-time}}, which itself relies on an adapted version of Allen's 13 temporal relations \cite{allen1981interval}.
\end{itemize}

\paragraph*{A Note on Temporal Identity}
Some ontological approaches assume a perdurantistic dogma, i.e.\ things are inherently four dimensional entities in space-time. 
\acs{AUTO} takes an endurantistic view: 
it specifies individuals in three dimensions and tracks temporal identity by the role $\mathtt{identical\_to}$. 


%% file: tikz/auto-architecture.tex
\begin{tikzpicture}
	
	\node[rectangle split, rectangle split parts=2, draw, text width=6.0cm, text centered] (tm) {\small \textbf{Traffic Models}\nodepart{second}\vspace{2.1cm}};

	\node[rectangle split, rectangle split parts=2, draw, text centered, text width=3.2cm, below right=0.7cm and 0.1cm of tm.north west] (stm) {\footnotesize \emph{Stationary\\Traffic Models} \nodepart{second}\vspace{0.6cm}};
	
	\node[rectangle, draw, below right=1.1cm and 0.1cm of stm.north west] (scenery) {\scriptsize Scenery};
	\node[rectangle, draw, right=0.1cm of scenery] (scene) {\scriptsize Scene};
	\node[rectangle, draw, right=0.1cm of scene] (situation) {\scriptsize Situation};
	
	\node[rectangle split, rectangle split parts=2, draw, text centered, text width=2.2cm, below left=0.7cm and 0.1cm of tm.north east] (etm) {\footnotesize \emph{Evolutionary Traffic Models} \nodepart{second}\vspace{0.6cm}};
	
	\node[rectangle, draw, below right=1.1cm and 0.1cm of etm.north west] (story) {\scriptsize Story};
	\node[rectangle, draw, right=0.1cm of story] (scenario) {\scriptsize Scenario};

	\node[rectangle split, rectangle split parts=2, draw, text width=2.0cm, text centered, below right=-0.8cm and 0.2cm of tm.north east] (related) {\small\textbf{Traffic Related}\nodepart{second}\vspace{3.3cm}};
	\node[rectangle, draw, below right=1.0cm and 0.1cm of related.north west] (time) {\footnotesize Time};
	\node[rectangle, draw, below right=0.1cm and 0cm of time.south west] (geo) {\footnotesize Geometry};
	\node[rectangle, draw, below right=0.1cm and 0cm of geo.south west] (phy) {\footnotesize Physics};
	\node[rectangle, draw, below right=0.1cm and 0cm of phy.south west] (act) {\footnotesize Act};
	\node[rectangle, draw, below right=0.1cm and 0cm of act.south west] (comm) {\footnotesize Communication};
	\node[rectangle, draw, below right=0.1cm and 0cm of comm.south west] (per) {\footnotesize Perception};
	
	\node[rectangle split, rectangle split parts=2, draw, text width=7.7cm, text centered, below right=1.05cm and -5.9cm of tm] (lm) {\small\textbf{6-Layer Model}\nodepart{second}\vspace{1.1cm}};
	\node[rectangle, draw, text width=0.95cm, below right=0.7cm and 0.1cm of lm.north west] (l1c) {\scriptsize L1 Core};
	\node[rectangle, draw, text width=0.95cm, below=0.15cm of l1c] (l1i) {\scriptsize L1 Impl.};
	
	\node[rectangle, draw, text width=0.95cm, right=0.1cm of l1c] (l2c) {\scriptsize L2 Core};
	\node[rectangle, draw, text width=0.95cm, below=0.15cm of l2c] (l2i) {\scriptsize L2 Impl.};
	
	\node[rectangle, draw, text width=0.95cm, right=0.1cm of l2c] (l3c) {\scriptsize L3 Core};
	\node[rectangle, draw, text width=0.95cm, below=0.15cm of l3c] (l3i) {\scriptsize L3 Impl.};
	
	\node[rectangle, draw, text width=0.95cm, right=0.1cm of l3c] (l4c) {\scriptsize L4 Core};
	\node[rectangle, draw, text width=0.95cm, below=0.15cm of l4c] (l4i) {\scriptsize L4 Impl.};
	
	\node[rectangle, draw, text width=0.95cm, right=0.1cm of l4c] (l5c) {\scriptsize L5 Core};
	\node[rectangle, draw, text width=0.95cm, below=0.15cm of l5c] (l5i) {\scriptsize L5 Impl.};
	
	\node[rectangle, draw, text width=0.95cm, right=0.1cm of l5c] (l6c) {\scriptsize L6 Core};
	\node[rectangle, draw, text width=0.95cm, below=0.15cm of l6c] (l6i) {\scriptsize L6 Impl.};
	
	\path[->] (tm.south) edge (tm.south |- lm.north);
	\path[->] (lm.north -| related.south) edge (related.south);
	
	\path[->] (tm.east) edge (tm.east -| related.west);
	\path[->] (l1i) edge (l1c);
	\path[->] (l2i) edge (l2c);
	\path[->] (l3i) edge (l3c);
	\path[->] (l4i) edge (l4c);
	\path[->] (l5i) edge (l5c);
	\path[->] (l6i) edge (l6c);
	
	\path[->] (scene) edge (scenery);
	
	\draw[->] (story) |- ++(0,-5mm) -| (situation);
	\draw[->] (scenario) |- ++(0,-6mm) -| (scene);
\end{tikzpicture}

%% file: sections/06_Formalization.tex

In general, criticality phenomena are not fixed to the combined DL-rule approach proposed in this work. 
Various approaches exhibit differences in ease and specificity of the modeling process, computational efficiency of decidability problems, and availability of expertise.
However, a common feature is their formalization being based on an implicit or explicit ontology. 
This section depicts the formalization process for our DL-rule approach as well as how the formalization can be leveraged for the construction of such an ontology. 

\subsection{The Formalization Process}
\label{subsec:formalization-proc}

Recall that the general method of \autoref{subsec:general-method} started the formalization process (step \Circled{1}) where a criticality phenomenon is formalized. 
We now explain this process in more detail, which is based on the previously introduced framework of \acs{AUTO} and a catalog\footnote{\url{https://github.com/lu-w/auto/blob/main/criticality\_phenomena.owl}} of criticality phenomena defined in natural language that was assembled during prior work \cite{neurohr2021criticality}. 
We illustrate the formalization of the exemplary criticality phenomenon \emph{'Bicyclist Illegitimately Riding Over Pedestrian Crossing'} (CP 69) as a special case of \emph{Misconduct} (CP 143). 

The first step involves the collection and assembly of knowledge regarding the semantics of the criticality phenomenon, which relies on standards, laws, and domain experts. 
For our example, we deem as necessary the semantics of the relevant regulatory aspects, bicyclists, pedestrian crossings, and riding over.
During the process, the stakeholders agree on
\begin{enumerate}
	\item the relevant regulatory aspects being given by the German traffic regulations (StVO) stating that a bicyclist can not exercise the right of way indicated by a pedestrian crossing if the bicyclist is mounted, 
	\item a bicyclist as a traffic participant on a bicycle, and
	\item a pedestrian crossing as a an area that is officially marked with corresponding white stripes.
	\item riding over as the movement of an actor across an area
\end{enumerate}

A suggestion of an (aspectual) semantics is then given by:

\emph{``A bicyclist that in one scene is not on a drivable lane, and in a directly succeeding scene on a pedestrian crossing where the bicyclist is exercising a right of way is using this pedestrian crossing illegitimately.''}

As temporal identities of traffic entities need to be tracked through different scenes, we express this using a rule-based approach such that individuals can be bound to variables: 
{\small
\begin{equation*}
	\begin{aligned}
	&\mathtt{Scene}(s_1) \wedge \mathtt{Scene}(s_2) \wedge \mathtt{has\_traffic\_entity}(s_1, b_1) \\&
	\wedge \mathtt{has\_traffic\_entity}(s_2, b_2) \wedge \mathtt{identical\_to}(b_2, b_1) \\&
	\wedge \sunderline{\mathtt{Bicyclist}}(b_1) \wedge \sunderline{\mathtt{Bicyclist}}(b_2) \wedge \mathtt{directly\_after}(s_2, s_1)
	\\&
	\wedge \sunderline{\mathtt{Non\_Driveable\_Lane}}(w) \wedge \mathtt{has\_traffic\_entity}(s_1, w) \\&
	\wedge \mathtt{intersects}(b_1, w) \wedge \sunderline{\mathtt{Pedestrian\_Crossing}}(c) \\&
	\wedge  \mathtt{has\_traffic\_entity}(s_2, c) \wedge \mathtt{intersects}(b_2, c) \\& \wedge \mathtt{Driveable\_Lane}(l) \wedge \mathtt{in\_scene}(l, s_2) \wedge \mathtt{intersects}(c, l) \\&
	\wedge \mathtt{Driver}(d) \wedge \mathtt{drives}(d, v) \wedge \mathtt{in\_scene}(d, s_2) \\&
	\wedge \mathtt{intersects}(d, l) \wedge \mathtt{has\_intersecting\_path}(v, b_2)\\&
	\rightarrow \mathtt{CP_{69}}(b_2)
	\end{aligned}
\end{equation*}
}

Intuitively, this rule states that if a bicyclist is driving on a non-drivable lane (e.g., sidewalk) in one scene and on a pedestrian crossing in a subsequent scene in which also a traffic participant on an intersecting lane with an intersecting path, then $\mathtt{CP_{69}}$ holds for that bicyclist. 
More specifically, the identity of the bicyclist in the first scene ($s_1$) is referred to as $b_1$, respectively as $b_2$ for the second scene ($s_2$). 
The scenes have $b_1$ and $b_2$ as their respective traffic entity (hence, $\mathtt{has\_traffic\_entity}$) and are directly after each other (hence, $\mathtt{directly\_after}$). 
Furthermore, $s_1$ has a non-drivable lane $w$ as its traffic entity which the bicyclist $b_1$ intersects. 
Similarly, the second scene $s_2$ has a pedestrian crossing $c$ and driveable lane $l$ that intersects $c$. Furthermore, $s_2$ has a driver $d$ within $s_2$ on lane $l$ that drives a vehicle $v$ which in turn has an intersecting path with the bicyclist $b_2$.

\begin{figure}[htb!]
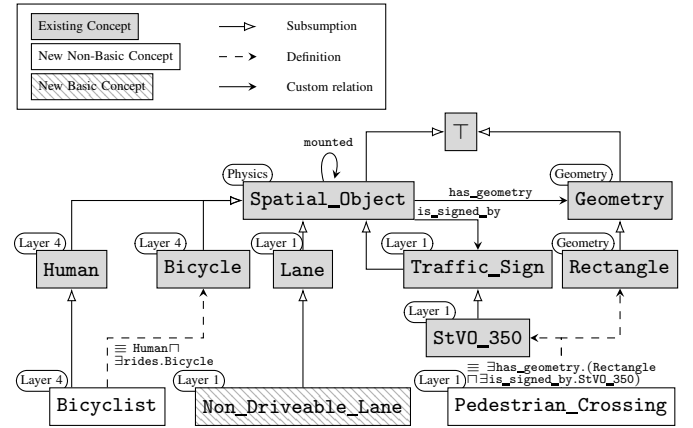

	\centering
	\include{tikz/cp-ex-subsumption}
	\caption{Derived concepts for the criticality phenomenon 'Bicyclist Illegitimately Riding Over Pedestrian Crossing' including their subsumption into relevant parts of the existing T-Box, as implemented in \acs{AUTO}}
	\label{fig:cp-ex-subsumption}
\end{figure}

The formalization relies on a set of concepts from \acs{AUTO} that are assumed to be already existent as well as suitably defined and axiomatized in the ontology, e.g.\ for temporal and spatial concepts. 
On the other hand, $\mathtt{Bicyclist}$, $\mathtt{Pedestrian\_Crossing}$, and $\mathtt{Non\_Driveable\_Lane}$ (underlined) are, for the scope of the example, not present at the time of formalization and therefore need to be added to $\mathcal{T}$. 
The following integration of concepts and their subsumption is also visualized in \autoref{fig:cp-ex-subsumption}. 

\begin{itemize}
	\item $\mathtt{Bicyclist} \equiv \mathtt{Human} \sqcap \exists \mathtt{rides} . \mathtt{Bicycle}$ 
	\begin{itemize}
		\item $\mathtt{Human}$, $\mathtt{Bicycle}$ are basic concepts
		\item $\mathtt{rides} \equiv \mathtt{mounted}$
		\begin{itemize}
			\item $\mathtt{mounted}$ is a basic concept (something physically and stably touches another spatial object)
		\end{itemize}
	\end{itemize}
	\item $\begin{aligned}[t]&\mathtt{Pedestrian\_Crossing} \equiv \exists \mathtt{has\_geometry}. \\&(\mathtt{Rectangle} \sqcap \exists \mathtt{is\_signed\_by}. \mathtt{StVO\_350})\end{aligned}$
	\begin{itemize}
		\item $\mathtt{has\_geometry}$ (geometries of spatial objects), $\mathtt{Rectangle}$ (a two-dimensional shape), $\mathtt{is\_signed\_by}$ (which traffic signs apply to an individual), and $\mathtt{StVO\_350}$ (marker for pedestrian crossing) are basic concepts
	\end{itemize}
	\item $\mathtt{Non\_Driveable\_Lane}$, a basic concept (a lane that is not intended to be driven on by any vehicle, e.g. a walkway)
\end{itemize}

Requiring that the necessary basic concepts can either be directly observed in the analyzed data or at least be derived therefrom, e.g.\ by means of machine learning algorithms such as image segmentation for $\mathtt{Non\_Driveable\_Lane}$, we now have the ability to (aspectually) infer the presence of $\mathtt{CP_{69}}$. 

\subsection{Approaching Formalizability}

Concepts and phenomena may be inherently unformalizable due to their vagueness (e.g. subjectivity and case-by-case decisions). 
To mitigate this problem, $\mathtt{CP_X}$ can be specified by:
\begin{enumerate}
	\item An \emph{exact representation:} if possible, we give an exact representation using an equivalence, i.e.\ $\mathtt{CP_X} \equiv f(\mathtt{C_k}, \dots, \mathtt{C_l})$ where $f$ is a logical concept expression.
	\item An \emph{under approximation}: if an exact representation is not possible, we use an approximation. In DL, an under approximation can be specified as a GCI of the form $\mathtt{CP_X} \sqsupseteq f(\mathtt{C_k}, \dots, \mathtt{C_l})$.
	On top of DL, sufficient conditions for $\mathtt{CP_X}$ can be stated as a rule, i.e.\ $\bigwedge_{r_a \in R_A} r_a \rightarrow \mathtt{CP_X}(\mathtt{v})$. 
	\item An \emph{over approximation}: an over approximation is specified by a GCI $\mathtt{CP_X} \sqsubseteq f(\mathtt{C_k}, \dots, \mathtt{C_l})$. 
	Combined with an under approximation, it allows for a two-sided bound.
\end{enumerate}

%

For example, 'Occluded Pedestrian' (CP 157) can be exactly formalized assuming that $\mathtt{Pedestrian}$, $\mathtt{Is\_Occlusion}$, and $\mathtt{is\_occluded}$ are exactly formalized or basic concepts: $\mathtt{CP_{157}} \equiv \mathtt{Is\_Occlusion} \sqcap \exists \mathtt{is\_occluded} . \mathtt{Pedestrian}$ 
The phenomenon 'Misconduct' (CP 143) on the other hand has a broad scope. 
We hence model certain aspects of it, such as driving without vehicle lights turned on in night scenarios: $\mathtt{CP_{143}} \sqsupseteq (\mathtt{Off\_Light\_Vehicle} \sqcap \exists \mathtt{in\_traffic\_model}.\mathtt{Night\_Scenario}) \sqcup \mathtt{CP_{69}}$ 
Similarly, 'Vulnerable Road User with Road Access' (CP 114), contains the vague concept Road Access, which was over approximated in terms of $\mathtt{is\_near}$. 
It is formalized as $\mathtt{CP_{114}} \sqsubseteq \mathtt{Vulnerable\_Road\_User} \sqcap \exists \mathtt{is\_near} . \mathtt{Driveable\_Lane}$. 
Such approximations can be iteratively refined.

\begin{table*}[htb]
	\caption{Exactly ($\equiv$) or over ($\sqsubseteq$) resp. under ($\sqsupseteq$) approximately formalized criticality phenomena for the prototypical evaluation.}
	\label{tab:cps}
	\footnotesize
	\centering
	\begin{tabularx}{\textwidth}{
			>{\raggedright\arraybackslash}p{0.08\textwidth}
			>{\raggedright\arraybackslash}p{0.11\textwidth}
			>{\raggedright\arraybackslash}p{0.19\textwidth}
			>{\raggedright\arraybackslash}p{0.15\textwidth}
			>{\raggedright\arraybackslash}p{0.11\textwidth}
			>{\raggedright\arraybackslash}p{0.12\textwidth}
			>{\raggedright\arraybackslash}p{0.1\textwidth}
		}
		\toprule
		\textbf{(L1) Road Network} & \textbf{(L2) Roadside Structures} & \textbf{(L4) Dynamic Objects} & \textbf{(L5) Environmental Conditions} & \textbf{Physics} & \textbf{Perception} & \textbf{Policy} \\
		\midrule
		Pedestrian Crossing / Ford$^\equiv$ & Building for Unpredictable Road Users near Road$^\sqsubseteq$ & Pedestrian on Roadway$^\equiv$\newline Vulnerable Road User with Road Access$^\sqsubseteq$\newline Passing of Parking Vehicle$^\equiv$\newline {Intersecting Planned Paths}$^\sqsupseteq$ & Heavy Rain$^\equiv$\newline Freezing Temperatures$^\equiv$ & {Small Distance}$^\sqsupseteq$\newline {Strong Braking Maneuver}$^\equiv$\newline {High Relative Speed}$^\equiv$ & {Occlusion}$^\equiv$\newline Occluded Pedestrian$^\equiv$\newline Occluded Traffic Infrastructure$^\equiv$ & Illegitimate Use of Pedestrian Crossing$^\sqsupseteq$\newline Misconduct$^\sqsupseteq$ \\
		\bottomrule
	\end{tabularx}
\end{table*}

Besides those three formalizations, our prototype contains the definitions of in total 16 criticality phenomena, depicted along their primary classification within \acs{AUTO} in \autoref{tab:cps}. 
They were selected from a catalog of roughly 300 criticality phenomena described in natural language which originated during an initial iteration of the criticality analysis. 
Their formalization is detailedly presented in the Appendix.

%% file: tikz/cp-ex-subsumption.tex
\footnotesize
\begin{tikzpicture}
	\node[draw, fill=gray!30, rectangle] at (-5, 1.4) (c1) {\tiny Existing Concept};
	\node[draw, fill=white, rectangle, below=0.4cm of c1.south west, anchor=south west] (c2) {\tiny New Non-Basic Concept};
	\node[draw, rectangle, pattern=north west lines, pattern color=gray!60, rectangle, below=0.4cm of c2.south west, anchor=south west] (c3) {\tiny New Basic Concept};
	\node[right=2.3cm of c1.west] (l1) {};
	\node[right=0.5cm of l1] (l2) {};
	\node[right=2.3cm of c2.west] (l3) {};
	\node[right=0.5cm of l3] (l4) {};
	\node[right=2.3cm of c3.west] (l5) {};
	\node[right=0.5cm of l5] (l6) {};
	\draw[-{Latex[open]}] (l1) -- (l2);
	\draw[-stealth,dashed] (l3) -- (l4);
	\draw[-stealth] (l5) -- (l6);
	\node[right=0.1cm of l2] (l2-l) {\tiny Subsumption};
	\node[right=0.1cm of l4] (l4-l) {\tiny Definition};
	\node[right=0.1cm of l6] (l6-l) {\tiny Custom relation};
	\draw[draw=black] (-5.9, 1.7) rectangle ++(4.9, -1.4);
	
	\node[minimum height=0.5cm, draw, fill=gray!30, rectangle] (top) {$\top$};
	\node[minimum height=0.5cm, draw, fill=gray!30, rectangle, below left=0.4cm and 0.4cm of top] (so)	{$\mathtt{Spatial\_Object}$};
	\node[minimum height=0.5cm, draw, fill=gray!30, rectangle, below left=0.4cm and -1.2cm of so] (lane)	{$\mathtt{Lane}$};
	\node[minimum height=0.5cm, draw, fill=gray!30, rectangle, below left=0.4cm and -0.1cm of so] (bic)	{$\mathtt{Bicycle}$};
	\node[minimum height=0.5cm, draw, fill=gray!30, rectangle, below left=0.4cm and 1.8cm of so] (human)	{$\mathtt{Human}$};
	\node[minimum height=0.5cm, draw, fill=gray!30, rectangle, below right=0.4cm and -0.15cm of so] (sign)	{$\mathtt{Traffic\_Sign}$};
	\node[minimum height=0.5cm, draw, fill=gray!30, rectangle, below=0.4cm of sign] (stvo)	{$\mathtt{StVO\_350}$};
	\node[minimum height=0.5cm, draw, fill=gray!30, rectangle, below right=0.4cm and 1.2cm of top] (geo)	{$\mathtt{Geometry}$};
	\node[minimum height=0.5cm, draw, fill=gray!30, rectangle, below=0.4cm of geo] (rect)	{$\mathtt{Rectangle}$};
	\node[minimum height=0.5cm, draw, rectangle, below right=0.4cm and -1.1cm of stvo, fill=white] (cross)	{$\mathtt{Pedestrian\_Crossing}$};
	\node[minimum height=0.5cm, draw, rectangle, fill=white] (bicyclist) at (human.east |- cross)	{$\mathtt{Bicyclist}$};
	\node[preaction={fill, white}, minimum height=0.5cm, draw, rectangle, pattern=north west lines, pattern color=gray!60] (ndlane) at (lane.north |- cross)	{$\mathtt{Non\_Driveable\_Lane}$};
	
	\draw[-{Latex[open]}] ([xshift=0.5cm] so.north) |- (top);
	\draw[-{Latex[open]}] (geo) |- (top);
	\path[-{Latex[open]}] (rect) edge (geo);
	\draw[-{Latex[open]}] (sign.west) -| ([xshift=0.5cm] so.south);
	\path[-{Latex[open]}] (lane.north) edge (so.south -| lane.north);
	\draw[-{Latex[open]}] (bic)  |- (so.west);
	\draw[-{Latex[open]}] (human) |- (so);
	\path[-{Latex[open]}] (stvo) edge (sign);
	\path[-{Latex[open]}] (ndlane) edge (lane);
	\draw[-stealth,dashed]  (cross) -- (stvo.east -| cross) -- (stvo.east);
	\draw[-stealth,dashed] (cross) -- (stvo.east -| cross) -| (rect.south);
	\node[above=0.0cm of cross,align=left,fill=white,inner sep=0] (f1) {\tiny$\equiv \exists \mathtt{has\_geometry}. (\mathtt{Rectangle} $\\[-0.15cm]\tiny$ \sqcap \exists \mathtt{is\_signed\_by}. \mathtt{StVO\_350})$};
	\draw[-stealth,dashed] (bicyclist) -- (stvo.east -| bicyclist) -| (bic);
	\node[above right=0.25cm and -0.7cm of bicyclist,align=left,fill=white,inner sep=0] (f1) {\tiny$\equiv \mathtt{Human} \sqcap$\\[-0.15cm]\tiny$\exists \mathtt{rides} . \mathtt{Bicycle}$};
	\path[-{Latex[open]}] (bicyclist.north -| human.south) edge (human);
	\path[-stealth] (so) edge node[above,yshift=-0.08cm] {\tiny$\mathtt{has\_geometry}$} (geo);
	\draw[-stealth] (so.south east) -| (sign.north);
	\node[above right=0.32cm and -1.9cm of sign] (signed)  {\tiny$\mathtt{is\_signed\_by}$};
	\path[] (so) edge[->,>={stealth},loop above] node[] {\tiny$\mathtt{mounted}$} (so);
	
	\begin{scope}[on background layer]
		\node[above left=-0.06 and -0.4cm of so,rectangle,fill=white, draw=black, rounded corners,inner sep=1.7pt] (sol) {\tiny Physics};
		\node[above left=-0.06 and -0.6cm of geo,rectangle,fill=white, draw=black, rounded corners,inner sep=1.7pt] (geol) {\tiny Geometry};
		\node[above left=-0.06 and -0.4cm of human,rectangle,fill=white, draw=black, rounded corners,inner sep=1.7pt] (humanl) {\tiny Layer 4};
		\node[above left=-0.06 and -0.4cm of bic,rectangle,fill=white, draw=black, rounded corners,inner sep=1.7pt] (bicl) {\tiny Layer 4};
		\node[above left=-0.06 and -0.25cm of bicyclist,rectangle,fill=white, draw=black, rounded corners,inner sep=1.7pt] (bicyclistl) {\tiny Layer 4};
		\node[above left=-0.06 and -0.25cm of cross,rectangle,fill=white, draw=black, rounded corners,inner sep=1.7pt] (crossl) {\tiny Layer 1};
		\node[above left=-0.06 and -0.4cm of ndlane, rectangle,fill=white, draw=black, rounded corners,inner sep=1.7pt] (ndlanel) {\tiny Layer 1};
		\node[above left=-0.06 and -0.4cm of lane, rectangle,fill=white, draw=black, rounded corners,inner sep=1.7pt] (lanel) {\tiny Layer 1};
		\node[above left=-0.06 and -0.4cm of sign, rectangle,fill=white, draw=black, rounded corners,inner sep=1.7pt] (signl) {\tiny Layer 1};
		\node[above left=-0.06 and -0.4cm of stvo, rectangle,fill=white, draw=black, rounded corners,inner sep=1.7pt] (stvol) {\tiny Layer 1};
		\node[above left=-0.06 and -0.7cm of rect, rectangle,fill=white, draw=black, rounded corners,inner sep=1.7pt] (rectl) {\tiny Geometry};
	\end{scope}
\end{tikzpicture}

%% file: sections/07_Evaluation.tex

We present a Python implementation\footnote{\url{https://github.com/lu-w/criticality-recognition}} of the process of \autoref{sec:criticality-reasoning} that is based on the ontology introduced in \autoref{sec:auto}. 
While the realization of step \Circled{1} has already been unveiled, we now focus on the data and reasoning steps, i.e.\ \Circled{2} and \Circled{3}. 

\subsection{The OMEGA-Format}
\label{subsec:omega}

As motivated in \autoref{sec:intro}, the use of various data sources is of high value when analyzing the criticality within traffic scenarios. 
Methodically, our approach rests upon a unified data format that allows to represent traffic scenarios from different sources, including drones, measurement vehicles, and stationary cameras.
This format therefore provides input for \acs{AUTO} A-Boxes. 
For this, the OMEGA-format was chosen \cite{scholtes22}. 
The data format is HDF5-based and designed to store reference data of urban automotive settings in an object-list structure together with map and weather information. 

\subsection{Ontology Mapping}
\label{subsec:onto-map}
In order for \ac{AUTO} to harness the information provided by OMEGA-compliant data sets, we need to map the OMEGA-format onto the concepts provided by \ac{AUTO}, i.e.\ a realization of step \Circled{2}. 
As the development of the OMEGA-format was performed in conjunction with the development of \ac{AUTO}, the format is naturally mappable to the concepts of the ontology.

For example, we translate each object of the OMEGA-type \texttt{reflector\_guidingLamps} into a corresponding axiom within the ABox. 
The following specification of the OMEGA-format defines reflector guiding lamps as
\begin{quotation}
	``6.1 Type: \texttt{reflector\_guidingLamps}: Any reflectors or lamps along the lane (reflectors and guiding lamps are assumed to be regularly placed, their exact position is not known.)''.
\end{quotation}
We create a new individual $\mathtt{i}$ as well as a corresponding boundary geometry $\mathtt{i_\mathtt{geom}}$ and map the definition to $\mathtt{i}$ by:
\begin{gather*}
	(\mathtt{System} \sqcap \forall \mathtt{consists\_of}. (\mathtt{Reflector\_Guiding\_Lamp}\\ \sqcap \exists \mathtt{within}. \{\mathtt{i_{geom}}\}))(\mathtt{i})
\end{gather*}


\subsection{Automated Recognition of Criticality Phenomena}
\label{subsec:ind-cps}
After creating an A-Box $\mathcal{A}$ based on $\mathcal{T}$ containing a set of $n$ scenarios, the na\"{i}ve option would be to run the reasoner directly on $\mathcal{O} = (\mathcal{T}, \mathcal{A})$. 
With this, two problems arise:
\begin{enumerate}
	\item deriving abstract facts from concrete domains, such as performing complex trigonometrical calculations to check whether one geometry occludes another, and
	\item exploding state spaces for large A-Boxes, especially when analyzing real world scenarios.
\end{enumerate}
We resolve both issues within our implementation. 

\paragraph{Deriving Abstract Facts from Concrete Domains}
Here, we propose an \emph{iterative augmentation approach} to lift A-Boxes from concrete to abstract domains. 
We provide a library where the user can easily specify definitions for arbitrary concepts $\mathtt{C}$ and (reified) roles $\mathtt{r}$ using Python code\footnote{\url{https://github.com/lu-w/owlready2-augmentator}}. 
This enables us to derive whether those concepts or roles are present by means of classical imperative programming. 
The library provides a high level interface that performs such an augmentation for each individual within $\mathcal{A}$ of an ontology $\mathcal{O}$, denoted by \Call{do\_augmentation}{$\mathcal{O}$}, changing $\mathcal{O}$ in-place and returning the number of performed augmentations. 

Finally, the lifting of domains needs to be integrated with the reasoner, whose interface is denoted by \Call{do\_reasoning}{$\mathcal{O}$}. 
This integration is sketched by \autoref{algo:augment-reason}. 

\alglanguage{pseudocode}
\begin{algorithm}
	\caption{Inference by lifting concrete to abstract domains}
	\label{algo:augment-reason}
	\textbf{Input:} an ontology $\mathcal{O}$ \\
	\textbf{Output:} $\mathcal{O}$ including all inferred facts 
	\begin{algorithmic}[1]
		\Procedure{Do\_Inference}{}
		\State $a = 1$
		\While{$a > 0$}
		\State $a =$ \Call{Do\_Augmentation}{$\mathcal{O}$}
		\State \Call{Do\_Reasoning}{$\mathcal{O}$}
		\EndWhile 
		\State \Return $\mathcal{O}$
		\EndProcedure
	\end{algorithmic}
\end{algorithm}
This simple yet correct procedure yields an iterative detection of all possible inferences on $\mathcal{O}$. 
Note that efficiency can be improved by directly integrating the augmentation into the reasoning engine, which was left for future work. 

\paragraph{Exploding State Spaces}
The second issue concerns the sheer size of real world scenarios. 
As an example, a complex urban traffic scene may contain 80 individuals on Layer 4 and 200 individuals on Layers 1 and 2. 
Even at a sampling rate of 3 Hertz, this amounts to over $2.5\cdot10^5$ individuals in 30 seconds and a combinatorial size of $6.25\cdot10^{10}$ and $1.5625\cdot10^{16}$ for potential binary and ternary relations. 

Upon closer inspection, many of those relations may not be logically possible. 
For example, spatial relations between a lane in scene one and a vehicle in scene two are irrelevant. 
Therefore, we safely use successive examination of the single scenes to correctly infer all non-temporal facts. 

This still leaves open the question of handling temporality. 
For example, the previously introduced rule for criticality phenomenon $\mathtt{CP_{69}}$ argues over two directly succeeding scenes, which can not be recognized on a scene level. 
However, running the reasoner directly on the scenario-level A-Box leads to the discussed scalability issues. 
Analogously to program slicing \cite{weiser1984program}, we suggest a temporal slicing approach.

\alglanguage{pseudocode}
\begin{algorithm}
	\caption{Temporal inference}
	\label{algo:temporal-reasoning}
	\textbf{Input:} an ontology $\mathcal{O} = (\mathcal{T}, \mathcal{A})$ with exactly one scenario $\mathtt{s}$ and scenes $\mathtt{s_1} \dots \mathtt{s_n}$ in $\mathcal{A}$ \\
	\textbf{Output:} $\mathcal{O}$ including inferred facts for $\mathtt{s}$
	\begin{algorithmic}[1]
		\Procedure{Do\_Temporal\_Reasoning}{}
		\For{$i \in \{1, \dots, n\}$}
			\State $\mathcal{A}_\mathtt{s_i}$\hspace{0.23cm}$ = $ \Call{Scene}{$\mathcal{A}, i$}
			\State \Call{Do\_Inference}{$(\mathcal{T}, \mathcal{A}_\mathtt{s_i})$}
		\EndFor
		\State $\eqmakebox[lhs1][l]{$\mathcal{A}_\mathit{merged}$} = $ \Call{Merge}{$\mathcal{A}_\mathtt{s_1}, \dots, \mathcal{A}_\mathtt{s_n}$}
		\State $\eqmakebox[lhs1][l]{$C$} = $ \Call{Temporal\_Concepts}{$\mathcal{T}$}
		\State $\eqmakebox[lhs1][l]{$\mathcal{A}_\mathit{sliced}$} = $ \Call{Slice}{$\mathcal{A}_\mathit{merged}, C$}
		\State \Call{Do\_Inference}{$(\mathcal{T}, \mathcal{A}_\mathit{sliced})$}
		\State $\eqmakebox[lhs1][l]{$\mathcal{A}_\mathit{res}$} = $ \Call{Merge}{$\mathcal{A}_\mathtt{merged}, \mathcal{A}_\mathtt{sliced}$}
		\State \Return $(\mathcal{T}, \mathcal{A}_\mathit{res})$
		\EndProcedure
	\end{algorithmic}
\end{algorithm}

Here, entities not relevant for temporal inference are sliced from the A-Box and re-added after finishing the temporal reasoning on the whole scenario, as shown in \autoref{algo:temporal-reasoning}. 
The algorithm relies on the following operators:
\begin{itemize}
	\item \Call{Scene}{$\mathcal{A}, i$}, which reduces $\mathcal{A}$ to an A-Box $\mathcal{A}_\mathtt{s_i}$ with only individuals of scene $\mathtt{s_i}$, i.e. $\mathcal{A}_\mathtt{s_i} \coloneqq \{\alpha \in \mathcal{A}\,|\, \exists \mathtt{n} \in \mathit{ind}(\alpha): (\exists \mathtt{has\_traffic\_entity}^{-1}.\{\mathtt{s_i}\})(\mathtt{n})\}$.
	\item \Call{merge}{$\mathcal{A}_1, \dots, \mathcal{A}_n$}, whose result is roughly equal to $\bigcup_{i\in \{1,\dots,n\}} \mathcal{A}_i$, but special temporally constant individuals (e.g.\ lanes) are not merged as to avoid duplicates. 
	Temporal identity is added by the \Call{merge}{$\cdot$} algorithm using the previously introduced role $\mathtt{identical\_to}$.
	\item \Call{Temporal\_Concepts}{$\mathcal{T}$}, which identifies a set $\mathsf{N}_T$ of concepts and roles that are (possibly indirectly) used in an axiom involving a concept from a set of temporal concepts $T$, e.g. $T = \{\mathtt{Interval}, \mathtt{Point}\}$. 
	Formally, $\mathsf{N}_T = \{\mathtt{C} \in \mathsf{N}_C \cup \mathsf{N}_R \,\mid\, \mathtt{D} \in T, k \geq 0, \exists \alpha_0 \in \mathit{axs}(\mathtt{C}), \exists \alpha_2, \dots, \alpha_{k-1} \in \mathcal{T}, \alpha_k \in \mathit{axs}(\mathtt{D}): \bigwedge_{i=0}^{k-1} (\mathit{nms}(\alpha_i) \cap \mathit{nms}(\alpha_{i+1}) \neq \emptyset) \}$. 
	In the case that $\mathcal{O}$ contains rules, $\mathsf{N}_T$ can be defined analogously.
	\item \Call{Slice}{$\mathcal{A}, \mathsf{T}$}, which creates an A-Box $\mathcal{A}_\mathit{sliced}$ with only individuals that relate to concepts in $\mathsf{T}$, i.e. $\mathcal{A}_\mathit{sliced} \coloneqq \{ \alpha \in \mathcal{A} \,|\, \exists \mathtt{i} \in \mathit{ind}(\alpha):\, \exists \mathtt{C} \in \mathsf{T}:\, \mathtt{C}(\mathtt{i}) \}$.
\end{itemize}
Intuitively, the algorithm takes as input an A-Box $\mathcal{A}$ with one scenario consisting of $n$ scenes. 
It creates an A-Box $\mathcal{A}_\mathit{res}$ where all temporal facts -- i.e. those on a scenario-level -- are inferred. 
This concerns, for example, the rule for $\mathtt{CP_{69}}$, which can only be applied when multiple scenes are present in the analyzed A-Box. 
Firstly, it infers all entailed facts on each scene using \autoref{algo:augment-reason}. 
Afterwards, it merges all scenes into a single A-Box $\mathcal{A}_\mathit{merged}$. 
Based on the formalized criticality phenomena in $\mathcal{T}$, it also derives a set of temporal concepts $C$ which are necessary to infer facts about the presence of temporal criticality phenomena. 
Those are then used to create a sliced A-Box $\mathcal{A}_\mathit{sliced}$ in which only individuals of concepts in $C$ remain. 
This sliced A-Box is then considerably smaller than the original A-Box and therefore suitable for the inference sketched in \autoref{algo:augment-reason}. 
Finally, all individuals that were previously sliced away are re-added to the A-Box $\mathcal{A}_\mathit{res}$. 

Since criticality phenomena have been modeled as axioms within $\mathcal{T}$, \autoref{algo:temporal-reasoning} eventually yields an inferred A-Box $\mathcal{A}$ in which the presence of criticality phenomena is documented. 

%% file: sections/08_Results.tex

We evaluate the implementation on the inD data set \cite{bock2019ind}, showing how the reasoner can answer the competency questions A1 to A3 based on the exemplary set of formalized phenomena and our implementation of \autoref{algo:augment-reason} and \autoref{algo:temporal-reasoning}. 
Albeit the main contribution of this work being the development of the method and tooling, we present a first evaluation for a semantic data analysis use case. 
Recall that we have previously argued that a verification and validation process relies on various recording modalities, including fleet monitoring, test vehicles, and drone data. 
As to exemplarily examine one of these relevant data sources, we have selected a large-scale drone data set which can act as a precursor to a rigorous investigation of a variety of data sources.  

\subsection{The inD data set}
\label{subsec:ind}
The inD data set consists of naturalistic trajectories of more than 11\,500 road users, including vehicles, bicyclists, and pedestrians, at four urban intersections in Aachen, Germany.  
The utilization of drones in the creation of the data set retains naturalistic driving behavior by being inconspicuous and aids the accuracy of positioning and object dimensions. 
Its distinctive perspective allows to draw conclusions about the behavior of traffic participants in certain situations -- e.g.\ bicyclists at pedestrian crossings -- which in turn can be constructively used as assumptions for designing, implementing, and safeguarding an ADS. 
Moreover, and in contrast to an in-vehicle perspective, we are able to assume a holistic view on the traffic scenario, i.e.\ tracking of 'behavior chains' where the decisions of traffic participants influences each other through transitive propagation. 
As an example of such a chain, consider an occluded pedestrian which leads to a strong braking maneuver of an intersecting vehicle that in turn leads to a potential rear-end collision of a following vehicle. 

\subsection{Reasoning Results}

In order to demonstrate the practicality and utility of our approach, we evaluate of the method on inD in three steps:

\begin{itemize}
	\item Firstly, we present an intuitive example depicting the inference results for a single, representative traffic conflict.
	\item Secondly, analyze four scenarios on each intersection.
	\item Lastly, we compare the inferences of our tooling against two criticality metrics as determined by prior work \cite{schneider2021towards}. 
\end{itemize}

\subsubsection{An Exemplary Traffic Conflict}

\begin{figure*}[htb!]
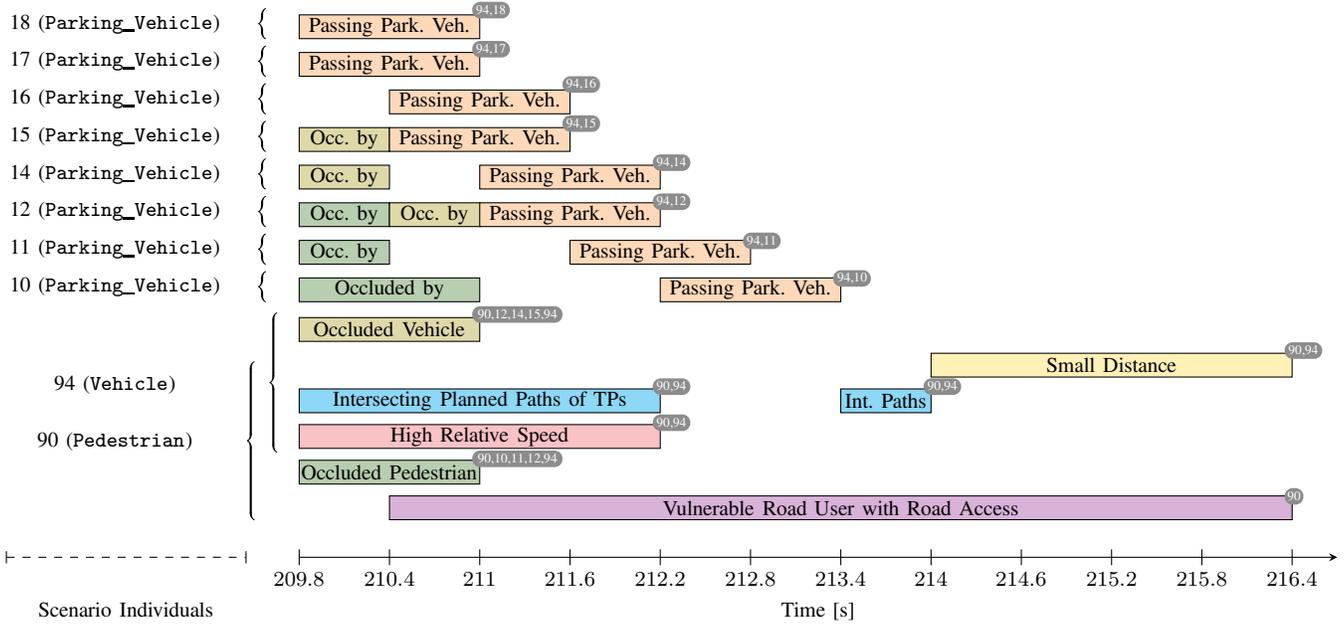

	\centering
	\include{tikz/eval_step1_scenario}
	\caption{The inferred criticality phenomena over time of a 6.6 second scenario with one moving vehicle, one pedestrian, and eight parking vehicles. The subject (and objects) are annotated by gray boxes on each criticality phenomenon. Occ. by = Occluded by and Passing Park. Veh. = Passing of Parking Vehicles.}
	\label{fig:eval-cps-conflict94}
\end{figure*}

We identified a simple yet representative traffic conflict within the inD data set\footnote{Namely, scenario number 94 in recording number 18}.  
Here, a passenger car approaches the intersection at the Frankenberger Park in Aachen, Germany, from the east while a pedestrian is occluded behind parking vehicles. 
The passenger car is forced to a substantial braking maneuver by the appearance of the pedestrian. 
The criticality phenomena within this conflict were inferred automatically by our tooling and are represented visually in \autoref{fig:eval-cps-conflict94}. 

The inferences show that the pedestrian and vehicle are indeed bilaterally occluded by parking vehicles while having a high relative speed during the first seconds of the scenario. 
Furthermore, the driver passes these parking vehicles while the pedestrian has access to the road. 
This leads to the occurrence of the more concrete criticality phenomenon Small Distance. 
Note that, albeit a harsh braking, the criticality phenomenon Strong Braking Maneuver was not detected, as the vehicle has been decelerated just below the threshold of $4.61$ m/s². 
This showcases an inherent limitation of the binary identification approach, where values just below thresholds are deemed irrelevant albeit possibly still influencing the overall criticality. 

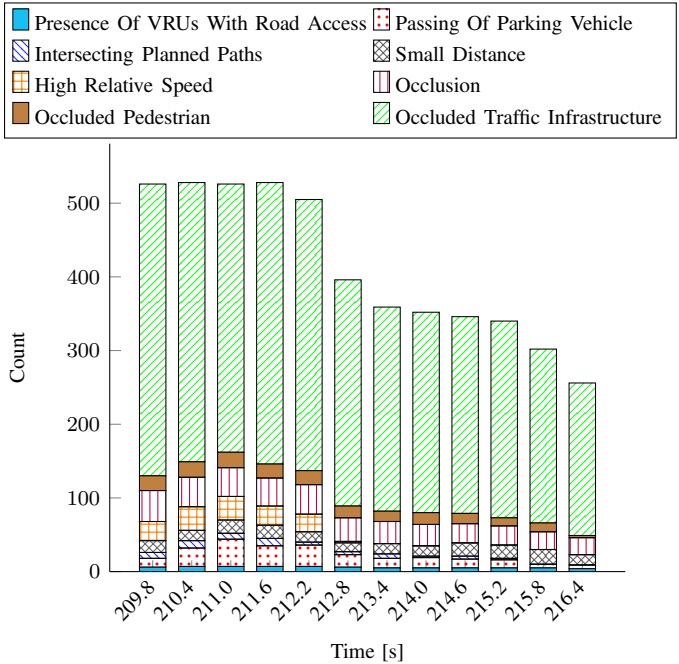
\begin{figure}[htb!]
	\centering
	\input{tikz/eval_step1_bar_chart.tikz}
	\caption{Identified criticality phenomena within the exemplary scenario.}
	\label{fig:all-cps-94}
\end{figure}

As an illustration of the aforementioned holistic view on traffic scenarios, \autoref{fig:all-cps-94} shows a bar chart of the amount of criticality phenomena per scene for the exemplary scenario. 

It is evident that occluded traffic infrastructures -- such as lane markings -- make up a large amount of the overall criticality phenomena. 
Furthermore, we find a considerable number of small distances and intersecting planned paths, as is to be expected in urban intersection scenarios. 
Overall, the amount of inferred criticality phenomena per scene is quite high, raising the issue of the specificity of the formalizations. 

\subsubsection{Evaluation of Multiple Scenarios}

We present an evaluation of in total 16 scenarios -- four scenarios for each of the four inD intersections -- with a mean duration of 11.27 $\pm$ 3.68 seconds covering a total span of 180.28 seconds with 361 unique traffic participants. 
The scenarios were converted into \acs{AUTO} A-Boxes at a rate of 1 Hertz, leading to 183 scenes. 
Due to the large scale of the data set, we reduced the evaluation to scene-level criticality phenomena, i.e.\ performing only \autoref{algo:augment-reason}. 
The overall number of inferred criticality phenomena amounts to 114\,922, resp. 628 per scene. 
If this number is again taken relative to the involved traffic participants, we find that there exist roughly 1.74 criticality phenomena per scene per traffic participant.

A detailed break-down of the inferred criticality phenomena per scene per traffic participant into the single phenomena classes for each of the intersections is presented in \autoref{tab:cp-per-tp-per-scene}. 

\begin{table}[htb!]
	\caption{Inferred criticality phenomena (CP) relative to the number of traffic participants (TP) and scenes over the selected 16 scenarios, arranged by intersection (Int.).}
	\label{tab:cp-per-tp-per-scene}
	\centering
	\scriptsize
	\begin{tabular}{lcccc}
		\toprule
		\textbf{CP per TP per Scene} & \textbf{Int. 1} & \textbf{Int. 2} & \textbf{Int. 3} & \textbf{Int. 4} \\
		\midrule
		High Relative Speed & 0.380 & 0.438 & 0.584 & 1.99 \\
		Intersecting Planned Paths & 0.058 & 0.066 & 0.050 & 0.047 \\
		Occluded Pedestrian & 0.811 & 1.332 & 0.377 & 0.291 \\
		Occluded Traffic Infrastructure & 21.781 & 13.127 & 47.658 & 32.076 \\
		Occlusion & 1.629 & 1.628 & 1.076 & 0.978 \\
		Presence of VRUs With Road Access & 0.257 & 0.180 & 0.094 & 0.191 \\
		Small Distance & 0.278 & 0.513 & 0.338 & 0.271 \\
		Strong Braking Maneuver & 0.003 & 0 & 0 & 0 \\
		\bottomrule
	\end{tabular}
\end{table}

For example, on intersection two, there exist 1.332 occluded pedestrians per traffic participant in each scene. 
The value is considerably lower for the other intersections, indicating a higher pedestrian occurrence or occlusion potential. 

To further show the capabilities of a semantic analysis, we demonstrate how a na\"ive analysis of the temporal relation of two criticality phenomena can be performed -- namely Intersecting Planned Paths and Occlusion -- on passenger car-pedestrian constellations. 
We first formulate two hypotheses: 
\begin{enumerate}
	\item[$H_1$] occlusions and intersecting paths are, independent of their temporal constellation, associated between passenger cars and pedestrians, and
	\item[$H_2$] in the cases where both occur, it is more likely to observe intersecting paths after occlusions than before. 
\end{enumerate}

We test $H_1$ by means of a simple contingency table as depicted in \autoref{tab:contingency}. 
This table displays the amount of cases of all four combinations of the occurrence of the two criticality phenomena between passenger cars and pedestrians. 

\begin{table}[htb!]
	\caption{Amount of the criticality phenomena for all pairwise combinations of passenger cars and pedestrians.}
	\label{tab:contingency}
	\scriptsize
	\centering
	\begin{tabular}{ccc}
		\toprule
		& \textbf{Occlusion} & \textbf{No Occlusion} \\
		\midrule
		\textbf{Intersecting Paths} & 7 & 1 \\
		\textbf{No Intersecting Paths} & 846 & 532 \\
		\bottomrule
	\end{tabular}
\end{table}

From this table, we compute the phi coefficient as $$\Phi = \frac{n_{11}n_{00}-n_{10}n_{01}}{\sqrt{(n_{10} + n_{11}) \cdot (n_{00} + n_{01}) \cdot (n_{00} + n_{01}) \cdot (n_{01} + n_{11})}}\text{,}$$
where $n_{ij}$ indicates the number $n$ in the $i$-th row and $j$-th column of the contingency table. 
For the analyzed criticality phenomena, this yields $\Phi = 0.04$. 
With the phi coefficient taking values from $[-1,1]$ in the dichotomous case, we find a minor yet positive correlation between the variables. 

As to test $H_2$, we first perform a simple analysis of the differences of the starting times of the intervals in which the criticality phenomena occur, taking into account all combinations of traffic participants. 
Since the association has been examined by $\Phi$, we can now focus on only the cases in which both small distances and occlusions occur. 
Our results indicate that, on average, small distances start 0.87 seconds after occlusions, with a standard deviation of 5.83 seconds. 


\subsubsection{Comparison against Criticality Metrics}

Besides comparing the relations between criticality phenomena themselves, one can also identify their relation to measured criticality as computed by criticality metrics. 
Here, our hypothesis states that an increase of a criticality metric is often preceded by a set of relevant criticality phenomena. 
This methodical approach obviously rests upon the assumption of a specific and sound formalization of criticality phenomena as well as upon employing valid criticality metrics. 

\begin{figure}[htb!]
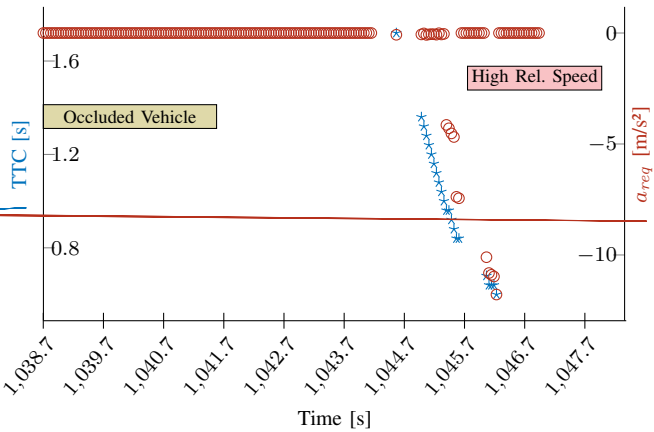

	\centering
	\include{tikz/eval_step3_scenario}
	\caption{The computed criticality metrics and inferred criticality phenomena for the exemplary scenario between both traffic participants over time.}
	\label{fig:eval-cps-conflict449}
\end{figure}

Prior work has already examined various criticality metrics for 22 traffic scenarios of inD for left-turn and passing maneuvers of vehicle pairs \cite{schneider2021towards}. 
We restrict our exemplary comparison to the analysis of one scenario and two metrics, namely scenario number 449 in recording number 28 and the Time To Collision (TTC) and Required Acceleration ($a_\mathit{req}$). 
Similar to \autoref{fig:eval-cps-conflict94}, the plot of this scenario is presented in \autoref{fig:eval-cps-conflict449} but now includes the metrics' values over time. 

In the example, a minimum TTC of 0.6 s and $a_\mathit{req}$ of -11.8 m/s² is given, with the metrics starting a linear approach of their minimum values at second 1044.7. 
After second 1046.7, the passing maneuver is finished. 
Our tooling observes that before this critical encounter, an occlusion took place between both traffic participants. 
Furthermore, we also identified high relative speeds during the passing phase of both vehicles. 

\paragraph*{A note on risk quantification} 
Within a rigorous safety case, this relation needs to be examined statistically on populations of representative scenarios, i.e.\ how strong is the association between certain phenomena and measured criticality across the OD. 
Depending on this strength, certain phenomena may be treated differently during the argumentation and evidence generation. 
For example, an argument as to why the ADS sufficiently mitigates the risk induced by occluded pedestrians may be elaborated on more extensively than those in which a lane marking is occluded -- depending on the associational findings and subsequent causal analyses. 
After this initial examination of the OD, safety standards require the development of a system design as well as a subsequent hazard analysis and assessment of risks. 
This assessment can be based upon prior knowledge that originated from the analysis of criticality metrics and phenomena. 
Since criticality phenomena represent factors that impact the safety in traffic in general, it is likely that such phenomena will also be relevant for the hazard analysis of the given system. 
The quantification provided by the aforementioned associational analysis of criticality can be used as an initial estimate for the risk assessment, although the specific system design may influence the actual risk. 
For example, an ADS could be designed in a way that effectively mitigates risks induced by occluded pedestrians. 
Nevertheless, an argument as to why this specific criticality phenomenon is handled sufficiently well by the system is mandatory. 


%% file: tikz/eval_step1_scenario.tex
\newcommand*{\TickSize}{3pt}
\newcommand*{\CPHeight}{9pt}
{\footnotesize
\begin{tikzpicture}
	\draw[draw=white] (-3.9, 1.35) rectangle (-1,1.75) node[pos=.5] {90 ($\mathtt{Pedestrian}$)};
	\draw[draw=white] (-3.9, 2.1) rectangle (-1,2.5) node[pos=.5] {94 ($\mathtt{Vehicle}$)};
	\draw[draw=white] (-3.9, 3.4) rectangle (-1,3.8) node[pos=.5] {10 ($\mathtt{Parking\_Vehicle}$)};
	\draw[draw=white] (-3.9, 3.9) rectangle (-1,4.3) node[pos=.5] {11 ($\mathtt{Parking\_Vehicle}$)};
	\draw[draw=white] (-3.9, 4.4) rectangle (-1,4.8) node[pos=.5] {12 ($\mathtt{Parking\_Vehicle}$)};
	\draw[draw=white] (-3.9, 4.9) rectangle (-1,5.3) node[pos=.5] {14 ($\mathtt{Parking\_Vehicle}$)};
	\draw[draw=white] (-3.9, 5.4) rectangle (-1,5.8) node[pos=.5] {15 ($\mathtt{Parking\_Vehicle}$)};
	\draw[draw=white] (-3.9, 5.9) rectangle (-1,6.3) node[pos=.5] {16 ($\mathtt{Parking\_Vehicle}$)};
	\draw[draw=white] (-3.9, 6.4) rectangle (-1,6.8) node[pos=.5] {17 ($\mathtt{Parking\_Vehicle}$)};
	\draw[draw=white] (-3.9, 6.9) rectangle (-1,7.3) node[pos=.5] {18 ($\mathtt{Parking\_Vehicle}$)};
	
	\draw [thin, |-|, dashed] (-3.9,0) --	node[below, yshift=-0.5cm] {Scenario Individuals}  (-0.7,0);
	\draw [thin, -stealth] (0,0) --	node[below, yshift=-0.5cm] {Time [s]}  ($(2*6.9,0)$);
	
	\foreach \x in {0, 0.6, 1.2, 1.8, 2.4, 3.0, 3.6, 4.2, 4.8, 5.4, 6.0, 6.6} {%
		\pgfmathsetmacro{\time}{\x+209.8}
		\draw ($(2*\x,0) + (0,-\TickSize)$) -- ($(2*\x,0) + (0,\TickSize)$)
		node [below, yshift=-0.2cm] {\pgfmathprintnumber[precision=1]{\time}};
	}

	\draw[thick, decorate, decoration = {calligraphic brace}] (-0.6,0.5) --  (-0.6,2.6);
	\draw[thick, decorate, decoration = {calligraphic brace}] (-0.3,1.4) --  (-0.3,3.25);
	\draw[thick, decorate, decoration = {calligraphic brace}] (-0.45,3.4) --  (-0.45,3.8);
	\draw[thick, decorate, decoration = {calligraphic brace}] (-0.45,3.9) --  (-0.45,4.3);
	\draw[thick, decorate, decoration = {calligraphic brace}] (-0.45,4.4) --  (-0.45,4.8);
	\draw[thick, decorate, decoration = {calligraphic brace}] (-0.45,4.9) --  (-0.45,5.3);
	\draw[thick, decorate, decoration = {calligraphic brace}] (-0.45,5.4) --  (-0.45,5.8);
	\draw[thick, decorate, decoration = {calligraphic brace}] (-0.45,5.9) --  (-0.45,6.3);
	\draw[thick, decorate, decoration = {calligraphic brace}] (-0.45,6.4) --  (-0.45,6.8);
	\draw[thick, decorate, decoration = {calligraphic brace}] (-0.45,6.9) --  (-0.45,7.3);

	\draw[draw=black,fill=violet!30] ($(2*0.6,0.5)$) rectangle ($(2*6.6,0.5) + (0,\CPHeight)$) node[pos=.5] {Vulnerable Road User with Road Access};

	\draw[draw=black,fill=OliveGreen!30] ($(2*0.0,0.5)+(0,1.5*\CPHeight)$) rectangle ($(2*1.2,0.5) + (0,2.5*\CPHeight)$) node[pos=.5] {Occluded Pedestrian};
	
	\draw[draw=black,fill=WildStrawberry!30] ($(2*0.0,0.5)+(0,3*\CPHeight)$) rectangle ($(4*1.2,0.5) + (0,4*\CPHeight)$) node[pos=.5] {High Relative Speed};
	
	\draw[draw=black,fill=Cyan!40] ($(2*0.0,0.5)+(0,4.5*\CPHeight)$) rectangle ($(4*1.2,0.5) + (0,5.5*\CPHeight)$) node[pos=.5] {Intersecting Planned Paths of TPs};
	
	\draw[draw=black,fill=Cyan!40] ($(6*1.2,0.5)+(0,4.5*\CPHeight)$) rectangle ($(7*1.2,0.5) + (0,5.5*\CPHeight)$) node[pos=.5] {Int. Paths};
	
	\draw[draw=black,fill=Goldenrod!40] ($(7*1.2,0.5)+(0,6*\CPHeight)$) rectangle ($(11*1.2,0.5) + (0,7*\CPHeight)$) node[pos=.5] {Small Distance};
	
	\draw[draw=black,fill=olive!30] ($(2*0.0,0.5)+(0,7.5*\CPHeight)$) rectangle ($(2*1.2,0.5) + (0,8.5*\CPHeight)$) node[pos=.5] {Occluded Vehicle};
	
	\draw[draw=black,fill=OliveGreen!30] ($(0*1.2,3.4)$) rectangle ($(2*1.2,3.4) + (0,\CPHeight)$) node[pos=.5] {Occluded by};
	\draw[draw=black,fill=OliveGreen!30] ($(0*1.2,3.9)$) rectangle ($(1*1.2,3.9) + (0,\CPHeight)$) node[pos=.5] {Occ. by};
	\draw[draw=black,fill=OliveGreen!30] ($(0*1.2,4.4)$) rectangle ($(1*1.2,4.4) + (0,\CPHeight)$) node[pos=.5] {Occ. by};
	
	\draw[draw=black,fill=olive!30] ($(1*1.2,4.4)$) rectangle ($(2*1.2,4.4) + (0,\CPHeight)$) node[pos=.5] {Occ. by};
	\draw[draw=black,fill=olive!30] ($(0*1.2,4.9)$) rectangle ($(1*1.2,4.9) + (0,\CPHeight)$) node[pos=.5] {Occ. by};
	\draw[draw=black,fill=olive!30] ($(0*1.2,5.4)$) rectangle ($(1*1.2,5.4) + (0,\CPHeight)$) node[pos=.5] {Occ. by};
	
	\draw[draw=black,fill=Apricot!50] ($(4*1.2,3.4)$) rectangle ($(6*1.2,3.4) + (0,\CPHeight)$) node[pos=.5] {Passing Park. Veh.};
	
	\draw[draw=black,fill=Apricot!50] ($(3*1.2,3.9)$) rectangle ($(5*1.2,3.9) + (0,\CPHeight)$) node[pos=.5] {Passing Park. Veh.};
	
	\draw[draw=black,fill=Apricot!50] ($(2*1.2,4.4)$) rectangle ($(4*1.2,4.4) + (0,\CPHeight)$) node[pos=.5] {Passing Park. Veh.};
	
	\draw[draw=black,fill=Apricot!50] ($(2*1.2,4.9)$) rectangle ($(4*1.2,4.9) + (0,\CPHeight)$) node[pos=.5] {Passing Park. Veh.};
	
	\draw[draw=black,fill=Apricot!50] ($(1*1.2,5.4)$) rectangle ($(3*1.2,5.4) + (0,\CPHeight)$) node[pos=.5] {Passing Park. Veh.};
	
	\draw[draw=black,fill=Apricot!50] ($(1*1.2,5.9)$) rectangle ($(3*1.2,5.9) + (0,\CPHeight)$) node[pos=.5] {Passing Park. Veh.};
	
	\draw[draw=black,fill=Apricot!50] ($(0*1.2,6.4)$) rectangle ($(2*1.2,6.4) + (0,\CPHeight)$) node[pos=.5] {Passing Park. Veh.};
	
	\draw[draw=black,fill=Apricot!50] ($(0*1.2,6.9)$) rectangle ($(2*1.2,6.9) + (0,\CPHeight)$) node[pos=.5] {Passing Park. Veh.};

	\fill[rounded corners=3pt, gray!90] (13.1,0.7) rectangle (13.35,0.92) node[pos=.5] {\tiny\color{white}90};
	\fill[rounded corners=3pt, gray!90] (2.3,1.2) rectangle (3.5,1.42) node[pos=.5] {\tiny\color{white}90,10,11,12,94};
	\fill[rounded corners=3pt, gray!90] (4.7,1.68) rectangle (5.2,1.9) node[pos=.5] {\tiny\color{white}90,94};
	\fill[rounded corners=3pt, gray!90] (4.7,2.16) rectangle (5.2,2.38) node[pos=.5] {\tiny\color{white}90,94};
	\fill[rounded corners=3pt, gray!90] (8.3,2.16) rectangle (8.8,2.38) node[pos=.5] {\tiny\color{white}90,94};
	\fill[rounded corners=3pt, gray!90] (13.1,2.64) rectangle (13.6,2.86) node[pos=.5] {\tiny\color{white}90,94};
	\fill[rounded corners=3pt, gray!90] (2.3,3.12) rectangle (3.5,3.34) node[pos=.5] {\tiny\color{white}90,12,14,15,94};
	\fill[rounded corners=3pt, gray!90] (7.1,3.6) rectangle (7.6,3.82) node[pos=.5] {\tiny\color{white}94,10};
	\fill[rounded corners=3pt, gray!90] (5.9,4.1) rectangle (6.4,4.32) node[pos=.5] {\tiny\color{white}94,11};
	\fill[rounded corners=3pt, gray!90] (4.7,4.62) rectangle (5.2,4.84) node[pos=.5] {\tiny\color{white}94,12};
	\fill[rounded corners=3pt, gray!90] (4.7,5.14) rectangle (5.2,5.36) node[pos=.5] {\tiny\color{white}94,14};
	\fill[rounded corners=3pt, gray!90] (3.5,5.66) rectangle (4,5.88) node[pos=.5] {\tiny\color{white}94,15};
	\fill[rounded corners=3pt, gray!90] (3.5,6.18) rectangle (4,6.4) node[pos=.5] {\tiny\color{white}94,16};
	\fill[rounded corners=3pt, gray!90] (2.3,6.65) rectangle (2.8,6.87) node[pos=.5] {\tiny\color{white}94,17};
	\fill[rounded corners=3pt, gray!90] (2.3,7.17) rectangle (2.8,7.39) node[pos=.5] {\tiny\color{white}94,18};
\end{tikzpicture}
}

%% file: tikz/eval_step1_bar_chart.tikz.tex
\begin{tikzpicture}
	\pgfplotstableread[col sep = semicolon]{ 
		Time [s];Presence Of VRUs With Road Access;Passing Of Parking Vehicle;Intersecting Planned Paths;Small Distance;High Relative Speed;Occlusion;Occluded Pedestrian;Occluded Traffic Infrastructure
		209.8;6;12;8;16;26;42;20;396
		210.4;7;25;10;14;32;40;21;379
		211.0;7;37;8;18;32;39;21;364
		211.6;7;28;10;18;26;38;19;382
		212.2;7;29;4;14;24;40;19;368
		212.8;6;17;4;12;2;32;16;307
		213.4;5;13;6;14;0;30;14;277
		214.0;5;14;2;14;0;29;16;272
		214.6;5;12;4;18;0;26;14;267
		215.2;5;11;2;18;0;26;11;267
		215.8;5;5;0;20;0;24;12;236
		216.4;4;5;0;14;0;23;3;207
	}\datatable
	
	\begin{axis}[
		ybar stacked,   
		ymin=0,         
		xtick=data,     
		xticklabels from table={\datatable}{Time [s]},
		x tick label style={rotate=45,anchor=east,yshift=-0.1cm},
		label style={font=\footnotesize},
		xlabel style={yshift=-0.3cm},
		tick label style={font=\footnotesize},
		ylabel={Count},
		xlabel={Time [s]},
		axis x line*=bottom,
		axis y line*=left,
		xtick pos=left,
		ytick pos=left,
		legend cell align={left},
		legend style={font=\footnotesize,
			at={(1.1,1.33)},
			anchor=north east,
			legend columns=2,
			row sep=0.005cm,
			column sep=0.005cm,
		}
		]
		\addplot [fill=cyan!70] table [x expr=\coordindex, y=Presence Of VRUs With Road Access] {\datatable};
		\addlegendentry{Presence Of VRUs With Road Access};
		\addplot [pattern color=red, pattern=dots] table [x expr=\coordindex, y=Passing Of Parking Vehicle] {\datatable};
		\addlegendentry{Passing Of Parking Vehicle};
		\addplot [pattern color=blue!80, pattern=north west lines] table [x expr=\coordindex, y=Intersecting Planned Paths] {\datatable};
		\addlegendentry{Intersecting Planned Paths};
		\addplot [pattern color=black!60, pattern=crosshatch] table [x expr=\coordindex, y=Small Distance] {\datatable};
		\addlegendentry{Small Distance};
		\addplot [pattern color=orange, pattern=grid] table [x expr=\coordindex, y=High Relative Speed] {\datatable};
		\addlegendentry{High Relative Speed};
		\addplot [pattern color=purple, pattern=vertical lines] table [x expr=\coordindex, y=Occlusion] {\datatable};
		\addlegendentry{Occlusion};
		\addplot [fill=brown] table [x expr=\coordindex, y=Occluded Pedestrian] {\datatable};
		\addlegendentry{Occluded Pedestrian};
		\addplot [pattern color=green!90, pattern=north east lines] table [x expr=\coordindex, y=Occluded Traffic Infrastructure] {\datatable};
		\addlegendentry{Occluded Traffic Infrastructure};
	\end{axis}
\end{tikzpicture}

%% file: tikz/eval_step3_scenario.tex
\newcommand*{\TickSize}{3pt}
\newcommand*{\CPHeight}{9pt}
\pgfplotsset{         
	width=1.051\columnwidth,     
	height=0.65\columnwidth 
}
{\footnotesize
\begin{tikzpicture}
	
	\draw [thin] (0,0) --	node[below, yshift=-1.1cm] {Time [s]}  ($(1.25*6.2,0)$);
	
	\foreach \x in {0, 1, 2, 3, 4, 5, 6, 7, 8, 9} {%
		\pgfmathsetmacro{\time}{\x+1038.68}
		\draw ($(0.8*\x,0) + (0,-\TickSize)$) -- ($(0.8*\x,0) + (0,\TickSize)$)
		node [below, yshift=-0.2cm,xshift=-0.17cm] {\rotatebox{50}{\pgfmathprintnumber[precision=1]{\time}}};
	}

	\draw[draw=black,fill=WildStrawberry!30] ($(0.8*7.05,2.5)+(0,1.8*\CPHeight)$) rectangle ($(0.8*9.28,2.5) + (0,2.8*\CPHeight)$) node[pos=.5] {\scriptsize High Rel. Speed};
%
%
%
%
%
	\draw[draw=black,fill=olive!30] ($(0.0,2.4)+(0,0.5*\CPHeight)$) rectangle ($(0.8*2.88,2.4) + (0,1.5*\CPHeight)$) node[pos=.5] {\scriptsize Occluded Vehicle};

	\begin{axis}[axis x line=none, yticklabel pos=right, ylabel={\color{RoyalBlue}\pgfuseplotmark{star} \hspace*{0.05cm} TTC [s]}, xmax=7.43, xmin=0, ticklabel style=right, axis y line*=left, ytick distance = 0.4, y label style={at={(axis description cs:0.12,0.5)}}]
		\addplot[RoyalBlue, only marks, mark=star] table[x=TimeRelPlot, y=TTC, col sep=semicolon] {tikz/metric.csv};
	\end{axis}

	\begin{axis}[axis x line=none, xmin=0, axis y line*=right, xmax=7.43, ylabel near ticks, yticklabel pos=right, ylabel={\color{BrickRed}{\pgfuseplotmark{o}} \hspace*{0.05cm} $a_\mathit{req}$ [m/s²]}, y label style={at={(axis description cs:1.0,0.5)}}, ticklabel style=left]
		\addplot[BrickRed, only marks, mark=o] table[x=TimeRelPlot, y=areq, col sep=semicolon] {tikz/metric.csv};
	\end{axis}
\end{tikzpicture}
}

%% file: sections/09_Conclusion.tex

We presented a first step towards a structured method to obtain a formalized definition of an OD, incorporated into the framework provided by the criticality analysis. 
This ontological OD definition has shown to be well-suited for reasoning over formalized criticality phenomena, thereby being a central element in a verification and validation strategy. 

Obviously, future work can address the extension of the knowledge base -- formalizing a wider variety of criticality phenomena --, the data sources -- analyzing data generated by test vehicles or simulation runs --, a rigorous data analysis -- studying the presence of (conjunctions of) criticality phenomena and their relation to traffic conflicts -- as well as a practical evaluation of the competency questions A4 and T1 to T3. 
An in-depth semantic analysis of large-scale intersection scenarios over a variety of criticality phenomena enables insights into the complex operating contexts of urban ADSs prior to system design. 
Furthermore, a first evaluation demonstrated the method's feasibility on a drone data set. However, its transferability to live settings under real-time constraints, e.g.\ for monitoring of ODD boundaries defined via criticality phenomena, is yet to be examined.
Incorporating a vehicle perspective (instead of the actor-agnostic view of drones) implicates the necessity of theoretically and practically addressing incomplete and uncertain knowledge about the environment originating from vehicle sensor data. 
Eventually, the results of this method can be used for safeguarding specific system designs. 
Future work has to address a rigorous analysis of the relation between criticality phenomena and criticality metrics as well as how these results can be used for the system's hazard analysis and risk quantification. 

Theoretically, we based the presented approach on DLs as well as rules and identified them as a suitable formalism to represent and reason on abstract knowledge about road traffic. 

For this theoretical aspect of the work, it remains an open question how to formalize knowledge about concrete domains directly in DLs. 
For the examined application of A-Box reasoning, we have tackled this challenge by means of augmentations that annotate existing spatio-temporal relations, but methodically, an integration into the logical framework is preferable. 
In the case of more abstract T-Box reasoning, it is not only preferable, but also mandatory to incorporate spatio-temporal semantics. 
For example, Allen's calculus states that $\mathtt{meets} \circ \mathtt{during} \sqsubseteq \mathtt{overlaps} \sqcup\mathtt{during} \sqcup \mathtt{starts}$, which is not a valid RIA in $\mathcal{SROIQ^{(D)}}$. 
In general, $\mathcal{SROIQ^{(D)}}$ has been shown not to be able to represent the semantics of such calculi \cite{hogenboom2010spatial}, i.e.\ relational composition tables, due to the lack of e.g. role negations. 
At least for RCC8, an alternative formalization using GCIs was proposed \cite{katz2005representing} and prototypically implemented \cite{stocker2009pelletspatial}, although the work seems to be abandoned. 
Therefore, theoretical advances need to address the incorporation of spatio-temporal semantics into the reasoning process, e.g. by integrating qualitative reasoning frameworks. 

We furthermore presented an implementation of the proposed method and evaluated a prototype on the large and multi-faceted A-Boxes of the inD drone data set. 

Practically, we found that reasoners, albeit mature, currently diminish in viability, leaving many of the available tools unmaintained. 
Another practical aspect concerns the extensive space and time consumption of the \textsc{Pellet} reasoner when running on the large A-Boxes of inD, and hence potential performance improvements by e.g.\ extending the idea of A-Box slicing. 
Alternatively, one refrain from using DL reasoners for inferring temporal properties. 
Scenarios can still be sliced, but analyzed with a solver for a scenario language. 

Lastly, the work has two assumptions: First, we assumed crisp terminological concepts within the T-Box. Second, the A-Box was considered perfectly representing ground truth. 

The first issue ignores that the OD model represents inherently fuzzy knowledge of humans -- e.g.\ 'heavy rain' may not have a crisp boundary. 
Representations of knowledge need to acknowledge vagueness within the T-Box and incorporate it into the reasoning process. 
The second issue concerns the applicability when presented with uncertain inputs. 
Here, one has to propagate those perception uncertainties along the inferences made by the reasoner. 
For a meaningful contribution to verification and validation, it is therefore imperative to both regard vagueness at design time and uncertainties at run time.

%% file: sections/10_Appendix.tex

We now display the formalization of the examined criticality phenomena as introduced in \autoref{tab:cps}. 
Employed concepts are only traced one layer down their recursive formalization. 
Basic concepts are marked by ${}^b$. 
Concepts that are augmented by an algebraic or first order logic formalization are indicated by $\coloneqq$ instead of the DL operators $\equiv$, $\sqsubseteq$, or $\sqsupseteq$. 
For this, we mix first-order predicate logic with DL and algebraic operations. 
Semantically, this is to be understood as a first order logic formula over the corresponding algebraic or geometric theory where the DL fragments, e.g. $\exists \mathtt{has\_traffic\_entity} . \{A_i\}$, represent sets of individuals on which we use set operators. 
These formulae are implemented using the augmentation library. 
We rely on some geometrical notation, listed in \autoref{tab:geo-symbol-notation}. 
We restrict the geometries to two dimension in Euclidean space and assume a single coordinate system. 

\begin{table}[htb!]
	\caption{Employed geometrical notation and functions.}
	\label{tab:geo-symbol-notation}
	\centering
	\footnotesize
	\begin{tabular}{ll}
		\toprule
		\textbf{Notation} & \textbf{Semantics} \\
		\midrule
		$p$ & A point in two-dimensional Euclidean space \\
		$p_i$ & The centroid's position of the geometry $g_i$ \\
		$d_i$ & The diameter of the geometry $g_i$ (m) \\
		\makecell[l]{$p_i^{fl}$, $p_i^{fr}$,\\ $p_i^{br}$, $p_i^{bl}$} & \makecell[l]{The front left, right and back left, right points of \\ the geometry $g_i$ (assuming a given yaw $\theta_i$)} \\
		$\text{dist}(p_1, p_2)$ & The euclidean distance between two points (m) \\
		$\text{area}(g_i)$ & The area of the geometry $g_i$ (m²) \\
		$\text{points}(g_i)$ & The set of points of the geometry $g_i$ \\
		$\text{ray}(p, \theta)$ & A ray (half-line) starting from $p$ with angle $\theta$ \\
		$\text{circle}(p, r)$ & A circle around $p$ with radius $r$ \\
		$\text{centroid}(g_i)$ & The centroid point of the geometry $g_i$ \\
		$\text{polygon}(P)$ & The polygon spanned by a tuple of points $P$ \\
		\bottomrule
	\end{tabular}
\end{table}

Overall, the algebraic and geometric formalizations are based on the notation depicted in \autoref{tab:math-symbol-notation}. 

\begin{table}[htb!]
	\caption{Employed mathematical identifiers including their corresponding DL representation (in \acs{AUTO}).}
	\label{tab:math-symbol-notation}
	\centering
	\footnotesize
	\begin{tabular}{p{0.8cm} p{3.6cm} p{3.1cm}}
		\toprule
		\textbf{Notation} & \textbf{Semantics} & \textbf{\acs{AUTO} representation} \\
		\midrule
		$A_i$ & A traffic participant with identifier no. $i$ & $\mathtt{Traffic\_Participant}(\mathtt{A})$, $(\exists \mathtt{identifier}^{-1} . \{A\})(i)$ \\
		$O$ & A spatial object & $\mathtt{Spatial\_Object}(O)$ \\
		$g_i$ & The geometry of $A_i$ described as a polygon (tuple of points) & $\mathtt{hasGeometry}$ \\
		$\theta_i$ & The yaw angle (°) of $A_i$ & $\mathtt{has\_yaw}$ \\
		$\theta_i^\mathit{max}$ & The assumed maximum yaw angle (°) of $A_i$ & $\mathtt{has\_maximum\_yaw}$ \\
		$\omega_i^\mathit{max}$ & The assumed maximum yaw rate (°/s) of $A_i$ & $\mathtt{has\_maximum\_yaw\_rate}$ \\
		$s_i$ & The speed (scalar, m/s) of $A_i$ & $\mathtt{has\_speed}$ \\
		$a_i$ & The acceleration (scalar, m/s²) of $A_i$ & $\mathtt{has\_acceleration}$ \\
		\bottomrule
	\end{tabular}
\end{table}

For an additional overview, \autoref{tab:target-values} presents a listing of all employed target values for criticality phenomena where thresholds have been applied or used within their concepts.

\begin{table}[htb!]
	\caption{Target values for each criticality phenomenon (CP).}
	\label{tab:target-values}
	\centering
	\footnotesize
	\begin{tabular}{cccc}
		\toprule
		\textbf{CP ID} & \textbf{Concept} & \textbf{Target Value(s)} & \textbf{Unit} \\
		\midrule
		113 & $\mathtt{is\_near}$ & 4 & m\\
		295 & $\mathtt{Heavy\_Rain}$ & 10 & mm/h\\
		295 & $\mathtt{Extremely\_Heavy\_Rain}$ & 50 & mm/h\\
		294 & $\mathtt{Freezing\_Air}$ & 0 & °C\\
		91 & $\mathtt{has\_intersecting\_path}$ & 8 ($\tau_1$) & s\\
		91 & $\mathtt{has\_intersecting\_path}$ & 3 ($\tau_2$) & s\\
		150 & $\text{relevant\_area}$ & 1 & s\\
		103 & $\mathtt{has\_acceleration}$ & -4.61, -3.3 & m/s²\\
		163 & $\mathtt{high\_relative\_speed}$ & 0.25 & -\\
		\bottomrule
	\end{tabular}
\end{table}

\paragraph{Pedestrian Crossing / Ford (CP 117)}

CP 117 is defined as the disjunction of pedestrian crossings and fords.
\begin{gather*}
	\mathtt{CP_{117}} \equiv \mathtt{Pedestrian\_Crossing}^b \sqcup \mathtt{Pedestrian\_Ford}^b
\end{gather*}

\paragraph{Building for Unpredictable Road Users near Road (CP 181)}

CP 181 is under-approximated by the (not exhaustive) union of retirement homes, schools, and kindergartens.
\begin{gather*}
	\mathtt{Kindergarten}^b \sqcup \mathtt{Retirement\_Home}^b \sqcup \mathtt{School}^b \sqsubseteq \mathtt{CP_{181}}
\end{gather*}

\paragraph{Pedestrian on Roadway (CP 82)}

CP 82 is a pedestrian that geometrically intersects with a driveable lane.
\begin{gather*}
	\mathtt{CP_{82}} \equiv \mathtt{Pedestrian} \sqcap \exists \mathtt{intersects} . \mathtt{Driveable\_Lane}^b
\end{gather*}
with 
$\mathtt{Pedestrian} \equiv \mathtt{Human} \sqcap \nexists \mathtt{drives} . \mathtt{Vehicle}$, and\\
$\mathtt{intersects}(a,b) \coloneqq \exists g_1 \in (\exists \mathtt{hasGeometry}^{-1} . \{a\}) \wedge \exists g_2 \in (\exists \mathtt{hasGeometry}^{-1} . \{b\}) . \text{points}(g_1) \cap \text{points}(g_2) \neq \emptyset$. 

\paragraph{Vulnerable Road User with Road Access (CP 113)}

CP 113 is a vulnerable road user near some driveable lane.
\begin{gather*}
	\mathtt{CP_{113}} \equiv \mathtt{VRU} \sqcap \exists \mathtt{is\_near} . \mathtt{Driveable\_Lane}^b
\end{gather*} 
with $\mathtt{VRU} \equiv \mathtt{Bicyclist} \sqcup \mathtt{Pedestrian}$, and 
$\mathtt{is\_near}(a,b) \coloneqq \text{dist}(a,b) < 4 \text{m}$.

\paragraph{Passing of Parking Vehicle (CP 91)}

CP 91 is a temporal CP and therefore defined as a passing activity that is conducted by a driver overtaking parking vehicles. 
\begin{gather*}
	\mathtt{CP_{91}} \equiv \mathtt{Pass} \sqcap \exists \mathtt{conducted\_by}^b . \mathtt{Driver} \sqcap \\
	\exists \mathtt{has\_participant}^b . \mathtt{Parking\_Vehicle}
\end{gather*}
with $\mathtt{Driver} \equiv \exists \mathtt{drives} . \mathtt{Vehicle}$ and
$\mathtt{Parking\_Vehicle} \sqsupseteq \mathtt{Vehicle} \sqcap \mathtt{Non\_Moving\_Dynamical\_Object} \sqcap \exists \mathtt{intersects} (\mathtt{Parking\_Lane \sqcup \mathtt{Parking\_Space}})$.
The passing activity is augmented using finite look back from the current scene $s_i$, searching for a scene $s_j$ and scenes $s_k$ between $s_i$ and $s_j$ satisfying the passing constraint. 
\begin{gather*}
	\mathtt{Pass}(A_i, O) \coloneqq  \\
	O \in (\mathtt{is\_in\_proximity}.\{A_i\}) \wedge O \in (\mathtt{behind}.\{A_i\}) \wedge \\
	\exists s_i \in (\exists \mathtt{has\_traffic\_entity} . \{A_i\}) . \\
	O \in (\exists \mathtt{in\_traffic\_model}.\{s_i\}) \wedge \exists s_j \in (\exists \mathtt{before} . \{s_i\}) . \\
	\exists A_j \in (\exists \mathtt{in\_traffic\_model}.\{s_j\} \sqcap \exists \mathtt{identical\_to}.\{A_i\}) . \\
	\exists O_j \in (\exists \mathtt{in\_traffic\_model}.\{s_j\} \sqcap \exists \mathtt{identical\_to}.\{O\}) . \\
	O_j \in (\exists \mathtt{is\_in\_proximity}.\{A_j\}) \wedge 
	O_j \in (\exists \mathtt{in\_front\_of}.\\\{A_j\}) 
	\forall s_k \in (\exists \mathtt{after} . \{s_j\} \sqcap \exists \mathtt{before} . \{s_i\}). \\
	\exists A_k \in (\exists \mathtt{in\_traffic\_model}.\{s_k\} \sqcap \exists \mathtt{identical\_to}.\{A_i\}) . \\
	\exists O_k \in (\exists \mathtt{in\_traffic\_model}.\{s_k\} \sqcap \exists \mathtt{identical\_to}.\{O\}) . \\
	O_k \in (\exists \mathtt{is\_in\_proximity}.\{A_k\}) \wedge \\
	O_k \in (\nexists \mathtt{in\_front\_of}.\{A_k\}) \wedge O_k \in (\nexists \mathtt{behind}.\{A_k\}) 
\end{gather*}
Due to mixing of universal, existential, and negated existential quantification, this formalization can not be depicted as a rule.

\paragraph{Intersecting Planned Paths (CP 293)}

Paths are represented as rays, for which we check whether they intersect in a given threshold ($\tau_1$) and how close it is ($\tau_2$). 
\begin{gather*}
	\mathtt{CP_{91}} \equiv \mathtt{Dynamical\_Object}^b \sqcap \\
	\exists \mathtt{has\_intersecting\_path} . \mathtt{Dynamical\_Object}^b
\end{gather*} with
\begin{gather*}
	\mathtt{has\_intersecting\_path}(A_1, A_2) \coloneqq \exists p \in \mathbb{R}^2. \\
	p \in \text{ray}(p_1, \theta_1) \cap \text{ray}(p_2, \theta_2) \wedge 
	\frac{\text{dist}(p_1, p)}{s_1} + \frac{\text{dist}(p_2, p)}{s_2} < \tau_1 \wedge \\
	\left|\frac{\text{dist}(p_1, p)}{s_1} - \frac{\text{dist}(p_2, p)}{s_2}\right| < \tau_2
\end{gather*}
where $\tau_1 \coloneqq 8 \text{ s}$ and $\tau_2 \coloneqq 3 \text{ s}$.

\paragraph{Heavy Rain (CP 295)}

CP 295 is defined as a rain with a precipitation intensity of at least 10 mm/h. 
\begin{gather*}
	\mathtt{CP_{295}} \equiv \mathtt{Extremely\_Heavy\_Rain} \sqcup \mathtt{Heavy\_Rain}
\end{gather*} with rules
$\mathtt{Rain}(r) \wedge \mathtt{has\_precipitation\_intensity}(r, x)$\\$\wedge x > 50 \rightarrow \mathtt{Extremely\_Heavy\_Rain}(r)$ and 
$\mathtt{Rain}(r) \wedge \mathtt{has\_precipitation\_intensity}(r, x) \wedge x \geq 10 \wedge x < 50 \rightarrow \mathtt{Heavy\_Rain}(r)$. 

\paragraph{Freezing Temperatures (CP 294)}

We define CP 294 as an air individual that has a temperature of below 0 °C. 
\begin{gather*}
	\mathtt{CP_{294}} \equiv \mathtt{Freezing\_Air}
\end{gather*} with
$\mathtt{Air}(a) \wedge \mathtt{has\_temperature}(a, t) \wedge t < 0 \rightarrow \mathtt{Freezing\_Air}(a)$.

\paragraph{Small Distance (CP 83)}

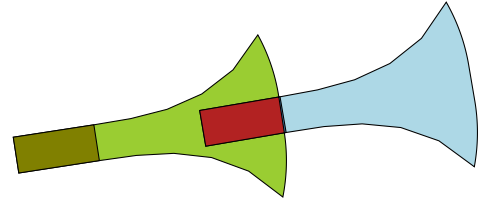
\begin{figure}[htb!]
	\centering
	\vspace{-0.6cm}
	\scalebox{0.4}{\input{pictures/sd_plt1.pgf}}
	\vspace{-0.5cm}
	\caption{The relevant areas of two vehicles in a following scenario.}
	\label{fig:small-dist}
\end{figure}

We identify any individual that has a small distance to some spatial object as CP 83. 
\begin{gather*}
	\mathtt{CP_{150}} \equiv \exists \mathtt{{object}_{CP_{150}}} . \mathtt{Spatial\_Object}
\end{gather*} with $\mathtt{{object}_{CP_{150}}}(A_1, A_2) \coloneqq \text{relevant\_area}(A_1) \cap \text{relevant\_area}(A_2) \neq \emptyset$. 
The relevant area is defined as the set of points that can be realistically reached:
\begin{gather*}
	\text{relevant\_area}(A_i) \coloneqq \{x \in \mathbb{R}^2 \mid \exists \omega \in [-\omega_i^\mathit{max}, \dots, \omega_i^\mathit{max}]. \\
	\exists t < t_H. \exists p \in \{p_i^{fl}, p_i^{fr}\}. \mathit{pos}_\omega(p, s_i, \theta_i, t) = x \}
\end{gather*} where 
$\mathit{pos}_\omega(p_i, s_i, \theta_i, t) \coloneqq p_i + s_i \cdot t \cdot (\cos (\theta_i^t), \sin (\theta_i^t))$ and\\
$\theta_i^t \coloneqq \begin{cases} (\theta_i + \text{sgn}(\omega) \cdot \theta_i^\mathit{max} \cdot t - \frac{{\theta_i^\mathit{max}}^2}{2 \cdot \omega}) & \text{if}\, |\omega \cdot t| > \theta_i^\mathit{max} \\ (\theta_i + \frac{\omega \cdot t^2}{2}) & \text{else} \end{cases}$ and $t_H$ is a time horizon (set to 1 s).
\autoref{fig:small-dist} shows two relevant areas and their intersection (in red). 

\paragraph{Strong Braking Maneuver (CP 103)}

Strong braking maneuvers are inferred based on the driven vehicle type. 
\begin{gather*}
	\mathtt{Motorized\_Road\_Vehicle}^b(v) \wedge \mathtt{has\_acceleration}^b(v, a) \\
	\wedge a < -4.61 \rightarrow \mathtt{CP_{103}}(v)\\
	\mathtt{Bicycle}^b(v) \wedge \mathtt{has\_acceleration}^b(v, a) \wedge a < -3.3 \\
	\rightarrow \mathtt{CP_{103}}(v)
\end{gather*}

\paragraph{High Relative Speed (CP 163)}

CP 163 is a vehicle with high relative speed, i.e. $\mathtt{CP_{163}} \equiv \exists \mathtt{{object}_{CP_{163}}} . \top$ with
$$
	\mathtt{object}_{\mathtt{CP_{163}}}(A_1, A_2) \coloneqq 
	\frac{||v_1-v_2||_2}{\min(s_\mathit{max}^\mathit{rule}, s_\mathit{max}^{A_1})} \geq 0.25
$$
where $s_\mathit{max}^\mathit{rule}$ is the maximum allowed speed and $s_\mathit{max}^{A_1}$ is the estimated maximum speed capability of $A_1$. 

\paragraph{Occlusion (CP 25)}

Occlusions are modeled as $n$-ary relations, allowing multiple occluding objects. 
CP 25 is the reified occlusion relation, i.e. $\mathtt{CP_{25}} \equiv \mathtt{Is\_Occlusion}$.

\begin{figure}[htb!]
	\centering
	\vspace*{-0.8cm}
	
	\hspace*{-1.22cm}
	\scalebox{0.25}{\input{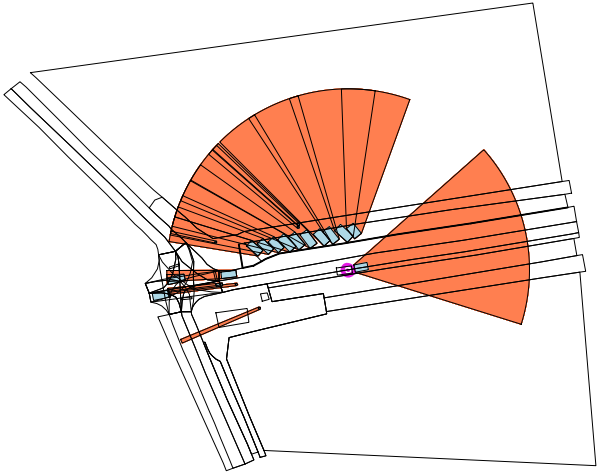}}
	\vspace*{-0.8cm}
	\caption{Visualization of occlusions on an inD scene. Occluded areas are orange, occluding objects are blue, and the observer is a purple dot.}
	\label{fig:occlusion}
\end{figure}

Our two-dimensional approach is visualized in \autoref{fig:occlusion}. 
We over approximate occlusions by assuming a circular field of view with some visibility range $r$ for an analyzed subject, i.e.\ $\text{fov}(A_i) \coloneqq \{x \in \mathbb{R}^2 \mid ||c_i-x|| \leq r \}$. 
$c_i$ is the point where the sensing of $A_i$ is assumed to be located. 
For simplicity, $$c_i \coloneqq \text{centroid}(g_i)  + (\cos(\theta_i), \sin(\theta_i)) \cdot \frac{||p_i^{bl} - p_i^{fl}||_2}{4}.$$

For a given opaque entity of non-zero height with its polygon exterior $g$ and a subject $A_i$, the left- and rightmost points and relative angles from the viewpoint of $c_i$ are:
\begin{itemize}
	\item $p_{g,i}^{l} \coloneqq \text{argmin}_{p \in g} \theta_\mathit{rel}(c_i, p)$
	\item $p_{g,i}^{r} \coloneqq \text{argmax}_{p \in g} \theta_\mathit{rel}(c_i, p)$
	\item $\theta_{g,i}^{l} \coloneqq \text{min}_{p \in g} \theta_\mathit{rel}(c_i, p)$
	\item $\theta_{g,i}^{r} \coloneqq \text{max}_{p \in g} \theta_\mathit{rel}(c_i, p)$
\end{itemize}
where $\theta_\mathit{rel}(c_i, p) \coloneqq \text{arctan2} (p_{i,1} - c_{i,1}, p_{i,0} - c_{i,0})$. 
For the edge case that $c_{i,1} \in [\text{min}_{p_i \in g} p_{i,1}, \text{max}_{p_i \in g} p_{i,1}]$ and $\exists p \in g. \theta_\mathit{rel}(c_i, p) \in [270, 360)$ and $\exists p \in g. \theta_\mathit{rel}(c_i, p) \in [0, 90)$, we note that min and max need to be switched in the definitions. 

The circular segment spanned by $g$ onto the boundary of the field of view is $\text{circular\_segment}(c_i, r, \theta_{g,i}^{l}, \theta_{g,i}^{r}) \coloneqq \{ r \cdot (\cos(\theta), \sin(\theta)) + c_i \mid \theta \in [\theta_{g,i}^{l}, \theta_{g,i}^{r}] \}.$ 
The exterior of $g$'s occluded area for $A_i$'s viewpoint is then $\text{occluded}(A_i, g) \coloneqq \text{polygon}(\text{circular\_segment}(c_i, r, \theta_{g,i}^{l}, \theta_{g,i}^{r}) \frown (p_{g,i}^{r}, p_{g,i}^{l})) \cup \text{polygon}(g)$, 
with $\frown$ denoting tuple concatenation. 
For a geometry $g_O$ of a spatial object $O$ and the subject actor $A_i$, we denote the rate of occlusion of $O$ through $g_1, \dots, g_n$ by\footnote{This is not perfectly accurate, but it suffices for our case.}
\begin{gather*}
	\delta(A_i, g_O, g_1, \dots, g_n) \coloneqq \\ \frac{\text{area}(g_O) \cap \bigcup_{j=1}^n \text{area}(\text{occluded}(A_i, g_j))}{\text{area}(g_O \cap \text{fov}(A_i))}\text{.}
\end{gather*}
If for an  $\mathtt{Observer}(A)$ and $\mathtt{Spatial\_Object}(O)$ it holds that there are $\{O_1, \dots, O_n\} \sqsubseteq (\mathtt{Spatial\_Object} \sqcap \neg \{A, O\})$ with $\delta(A, O, O_1, \dots, O_n) = d > 0 \wedge \bigwedge_{k=1}^n \exists h \in (\exists \mathtt{has\_height}^{-1}. \{O_k\}). h > 0$
then it there exists an individual $\mathtt{Is\_Occlusion}(N)$ with 
\begin{itemize}
	\item $\mathtt{is\_occluded\_for}(N, A)$,
	\item $\mathtt{is\_occluded\_by}(N, O_1), \dots, \mathtt{is\_occluded\_by}(N, O_n)$,
	\item $\mathtt{is\_occluded}(N, O)$,
	\item $\mathtt{has\_occlusion\_rate}(N, d)$.
\end{itemize}

\paragraph{Occluded Pedestrian (CP 160)}

This phenomenon is: 
\begin{gather*}
	\mathtt{CP_{160}} \equiv \mathtt{Is\_Occlusion} \sqcap \exists \mathtt{is\_occluded} . \mathtt{Pedestrian}
\end{gather*}

\paragraph{Occluded Traffic Infrastructure (CP 265)}

Similarly: 
\begin{gather*}
\mathtt{CP_{265}} \equiv \mathtt{Is\_Occlusion} \sqcap \\ \exists \mathtt{is\_occluded} . \mathtt{Traffic\_Infrastrcture}
\end{gather*}

\paragraph{Illegitimate Use of Pedestrian Crossing (CP 69)}

The formalizing rule is detailedly presented in \autoref{subsec:formalization-proc}.

\paragraph{Misconduct (CP 143)}

We base CP 143 on CP 69 as well as not turning on the vehicle lights in night scenarios. 
\begin{gather*}
	\mathtt{CP_{143}} \sqsupseteq (\mathtt{Off\_Light\_Vehicle} \sqcap \\
	\exists \mathtt{in\_traffic\_model}^b. \mathtt{Night\_Scenario}) \sqcup \mathtt{CP_{69}}
\end{gather*} with
$\mathtt{Off\_Light\_Vehicle} \equiv \mathtt{Vehicle} \sqcap 	\nexists \mathtt{consists\_of} (\mathtt{Headlight \sqcap \mathtt{Active\_Lamp}})$ and $\mathtt{Night\_Scenario} \equiv \mathtt{Scenario} \sqcap \exists \mathtt{has\_traffic\_entity} . \mathtt{Night\_Environment}\text{.}$

%% file: pictures/sd_plt1.pgf
\begingroup%
\makeatletter%
\begin{pgfpicture}%
\pgfpathrectangle{\pgfpointorigin}{\pgfqpoint{8.620000in}{4.490000in}}%
\pgfusepath{use as bounding box, clip}%
\begin{pgfscope}%
\pgfsetbuttcap%
\pgfsetmiterjoin%
\definecolor{currentfill}{rgb}{1.000000,1.000000,1.000000}%
\pgfsetfillcolor{currentfill}%
\pgfsetlinewidth{0.000000pt}%
\definecolor{currentstroke}{rgb}{1.000000,1.000000,1.000000}%
\pgfsetstrokecolor{currentstroke}%
\pgfsetdash{}{0pt}%
\pgfpathmoveto{\pgfqpoint{0.000000in}{0.000000in}}%
\pgfpathlineto{\pgfqpoint{8.620000in}{0.000000in}}%
\pgfpathlineto{\pgfqpoint{8.620000in}{4.490000in}}%
\pgfpathlineto{\pgfqpoint{0.000000in}{4.490000in}}%
\pgfpathclose%
\pgfusepath{fill}%
\end{pgfscope}%
\begin{pgfscope}%
\pgfpathrectangle{\pgfqpoint{1.077500in}{0.819137in}}{\pgfqpoint{6.680500in}{2.806826in}}%
\pgfusepath{clip}%
\pgfsetbuttcap%
\pgfsetmiterjoin%
\definecolor{currentfill}{rgb}{0.678431,0.847059,0.901961}%
\pgfsetfillcolor{currentfill}%
\pgfsetlinewidth{1.003750pt}%
\definecolor{currentstroke}{rgb}{0.000000,0.000000,0.000000}%
\pgfsetstrokecolor{currentstroke}%
\pgfsetdash{}{0pt}%
\pgfpathmoveto{\pgfqpoint{4.871492in}{2.259064in}}%
\pgfpathlineto{\pgfqpoint{5.363783in}{2.350092in}}%
\pgfpathlineto{\pgfqpoint{5.847586in}{2.482191in}}%
\pgfpathlineto{\pgfqpoint{6.308720in}{2.695073in}}%
\pgfpathlineto{\pgfqpoint{6.721920in}{3.024633in}}%
\pgfpathlineto{\pgfqpoint{7.046349in}{3.498380in}}%
\pgfpathlineto{\pgfqpoint{7.057081in}{3.479354in}}%
\pgfpathlineto{\pgfqpoint{7.067647in}{3.460235in}}%
\pgfpathlineto{\pgfqpoint{7.078046in}{3.441024in}}%
\pgfpathlineto{\pgfqpoint{7.088276in}{3.421723in}}%
\pgfpathlineto{\pgfqpoint{7.098337in}{3.402334in}}%
\pgfpathlineto{\pgfqpoint{7.108229in}{3.382858in}}%
\pgfpathlineto{\pgfqpoint{7.117951in}{3.363296in}}%
\pgfpathlineto{\pgfqpoint{7.127502in}{3.343651in}}%
\pgfpathlineto{\pgfqpoint{7.136881in}{3.323922in}}%
\pgfpathlineto{\pgfqpoint{7.146087in}{3.304113in}}%
\pgfpathlineto{\pgfqpoint{7.155120in}{3.284223in}}%
\pgfpathlineto{\pgfqpoint{7.163979in}{3.264256in}}%
\pgfpathlineto{\pgfqpoint{7.172664in}{3.244212in}}%
\pgfpathlineto{\pgfqpoint{7.181173in}{3.224094in}}%
\pgfpathlineto{\pgfqpoint{7.189506in}{3.203901in}}%
\pgfpathlineto{\pgfqpoint{7.197663in}{3.183637in}}%
\pgfpathlineto{\pgfqpoint{7.205643in}{3.163303in}}%
\pgfpathlineto{\pgfqpoint{7.213445in}{3.142899in}}%
\pgfpathlineto{\pgfqpoint{7.221069in}{3.122428in}}%
\pgfpathlineto{\pgfqpoint{7.228513in}{3.101892in}}%
\pgfpathlineto{\pgfqpoint{7.235779in}{3.081291in}}%
\pgfpathlineto{\pgfqpoint{7.242864in}{3.060628in}}%
\pgfpathlineto{\pgfqpoint{7.249768in}{3.039903in}}%
\pgfpathlineto{\pgfqpoint{7.256492in}{3.019120in}}%
\pgfpathlineto{\pgfqpoint{7.263034in}{2.998278in}}%
\pgfpathlineto{\pgfqpoint{7.269393in}{2.977380in}}%
\pgfpathlineto{\pgfqpoint{7.275570in}{2.956427in}}%
\pgfpathlineto{\pgfqpoint{7.281564in}{2.935421in}}%
\pgfpathlineto{\pgfqpoint{7.287375in}{2.914364in}}%
\pgfpathlineto{\pgfqpoint{7.293001in}{2.893257in}}%
\pgfpathlineto{\pgfqpoint{7.298444in}{2.872101in}}%
\pgfpathlineto{\pgfqpoint{7.303701in}{2.850899in}}%
\pgfpathlineto{\pgfqpoint{7.308773in}{2.829652in}}%
\pgfpathlineto{\pgfqpoint{7.313659in}{2.808361in}}%
\pgfpathlineto{\pgfqpoint{7.318360in}{2.787028in}}%
\pgfpathlineto{\pgfqpoint{7.322874in}{2.765655in}}%
\pgfpathlineto{\pgfqpoint{7.327201in}{2.744244in}}%
\pgfpathlineto{\pgfqpoint{7.331342in}{2.722796in}}%
\pgfpathlineto{\pgfqpoint{7.335295in}{2.701312in}}%
\pgfpathlineto{\pgfqpoint{7.339060in}{2.679795in}}%
\pgfpathlineto{\pgfqpoint{7.422319in}{2.190920in}}%
\pgfpathlineto{\pgfqpoint{7.425708in}{2.169340in}}%
\pgfpathlineto{\pgfqpoint{7.428909in}{2.147732in}}%
\pgfpathlineto{\pgfqpoint{7.431922in}{2.126096in}}%
\pgfpathlineto{\pgfqpoint{7.434745in}{2.104435in}}%
\pgfpathlineto{\pgfqpoint{7.437379in}{2.082750in}}%
\pgfpathlineto{\pgfqpoint{7.439824in}{2.061043in}}%
\pgfpathlineto{\pgfqpoint{7.442079in}{2.039316in}}%
\pgfpathlineto{\pgfqpoint{7.444145in}{2.017569in}}%
\pgfpathlineto{\pgfqpoint{7.446021in}{1.995806in}}%
\pgfpathlineto{\pgfqpoint{7.447707in}{1.974027in}}%
\pgfpathlineto{\pgfqpoint{7.449202in}{1.952234in}}%
\pgfpathlineto{\pgfqpoint{7.450508in}{1.930428in}}%
\pgfpathlineto{\pgfqpoint{7.451623in}{1.908613in}}%
\pgfpathlineto{\pgfqpoint{7.452547in}{1.886788in}}%
\pgfpathlineto{\pgfqpoint{7.453282in}{1.864956in}}%
\pgfpathlineto{\pgfqpoint{7.453825in}{1.843118in}}%
\pgfpathlineto{\pgfqpoint{7.454178in}{1.821277in}}%
\pgfpathlineto{\pgfqpoint{7.454341in}{1.799433in}}%
\pgfpathlineto{\pgfqpoint{7.454313in}{1.777589in}}%
\pgfpathlineto{\pgfqpoint{7.454094in}{1.755746in}}%
\pgfpathlineto{\pgfqpoint{7.453685in}{1.733905in}}%
\pgfpathlineto{\pgfqpoint{7.453085in}{1.712069in}}%
\pgfpathlineto{\pgfqpoint{7.452294in}{1.690239in}}%
\pgfpathlineto{\pgfqpoint{7.451313in}{1.668417in}}%
\pgfpathlineto{\pgfqpoint{7.450142in}{1.646604in}}%
\pgfpathlineto{\pgfqpoint{7.448780in}{1.624802in}}%
\pgfpathlineto{\pgfqpoint{7.447228in}{1.603013in}}%
\pgfpathlineto{\pgfqpoint{7.445486in}{1.581238in}}%
\pgfpathlineto{\pgfqpoint{7.443554in}{1.559480in}}%
\pgfpathlineto{\pgfqpoint{7.441433in}{1.537739in}}%
\pgfpathlineto{\pgfqpoint{7.439121in}{1.516017in}}%
\pgfpathlineto{\pgfqpoint{7.436620in}{1.494316in}}%
\pgfpathlineto{\pgfqpoint{7.433930in}{1.472638in}}%
\pgfpathlineto{\pgfqpoint{7.431051in}{1.450985in}}%
\pgfpathlineto{\pgfqpoint{7.427983in}{1.429357in}}%
\pgfpathlineto{\pgfqpoint{7.424727in}{1.407757in}}%
\pgfpathlineto{\pgfqpoint{7.421281in}{1.386186in}}%
\pgfpathlineto{\pgfqpoint{7.417648in}{1.364646in}}%
\pgfpathlineto{\pgfqpoint{7.413827in}{1.343138in}}%
\pgfpathlineto{\pgfqpoint{6.950741in}{1.682610in}}%
\pgfpathlineto{\pgfqpoint{6.451678in}{1.856625in}}%
\pgfpathlineto{\pgfqpoint{5.946056in}{1.904671in}}%
\pgfpathlineto{\pgfqpoint{5.445813in}{1.868986in}}%
\pgfpathlineto{\pgfqpoint{4.951173in}{1.791738in}}%
\pgfpathlineto{\pgfqpoint{3.899691in}{1.612456in}}%
\pgfpathlineto{\pgfqpoint{3.820010in}{2.079781in}}%
\pgfpathclose%
\pgfusepath{stroke,fill}%
\end{pgfscope}%
\begin{pgfscope}%
\pgfpathrectangle{\pgfqpoint{1.077500in}{0.819137in}}{\pgfqpoint{6.680500in}{2.806826in}}%
\pgfusepath{clip}%
\pgfsetbuttcap%
\pgfsetmiterjoin%
\definecolor{currentfill}{rgb}{0.603922,0.803922,0.196078}%
\pgfsetfillcolor{currentfill}%
\pgfsetlinewidth{1.003750pt}%
\definecolor{currentstroke}{rgb}{0.000000,0.000000,0.000000}%
\pgfsetstrokecolor{currentstroke}%
\pgfsetdash{}{0pt}%
\pgfpathmoveto{\pgfqpoint{2.435363in}{1.892158in}}%
\pgfpathlineto{\pgfqpoint{2.918048in}{1.973482in}}%
\pgfpathlineto{\pgfqpoint{3.393075in}{2.095090in}}%
\pgfpathlineto{\pgfqpoint{3.847198in}{2.296024in}}%
\pgfpathlineto{\pgfqpoint{4.256278in}{2.611771in}}%
\pgfpathlineto{\pgfqpoint{4.580820in}{3.069858in}}%
\pgfpathlineto{\pgfqpoint{4.591015in}{3.051090in}}%
\pgfpathlineto{\pgfqpoint{4.601047in}{3.032235in}}%
\pgfpathlineto{\pgfqpoint{4.610913in}{3.013293in}}%
\pgfpathlineto{\pgfqpoint{4.620614in}{2.994265in}}%
\pgfpathlineto{\pgfqpoint{4.630148in}{2.975153in}}%
\pgfpathlineto{\pgfqpoint{4.639516in}{2.955959in}}%
\pgfpathlineto{\pgfqpoint{4.648715in}{2.936684in}}%
\pgfpathlineto{\pgfqpoint{4.657746in}{2.917329in}}%
\pgfpathlineto{\pgfqpoint{4.666607in}{2.897897in}}%
\pgfpathlineto{\pgfqpoint{4.675299in}{2.878387in}}%
\pgfpathlineto{\pgfqpoint{4.683820in}{2.858803in}}%
\pgfpathlineto{\pgfqpoint{4.692170in}{2.839145in}}%
\pgfpathlineto{\pgfqpoint{4.700348in}{2.819415in}}%
\pgfpathlineto{\pgfqpoint{4.708353in}{2.799614in}}%
\pgfpathlineto{\pgfqpoint{4.716186in}{2.779744in}}%
\pgfpathlineto{\pgfqpoint{4.723845in}{2.759807in}}%
\pgfpathlineto{\pgfqpoint{4.731329in}{2.739803in}}%
\pgfpathlineto{\pgfqpoint{4.738639in}{2.719735in}}%
\pgfpathlineto{\pgfqpoint{4.745773in}{2.699604in}}%
\pgfpathlineto{\pgfqpoint{4.752731in}{2.679411in}}%
\pgfpathlineto{\pgfqpoint{4.759513in}{2.659159in}}%
\pgfpathlineto{\pgfqpoint{4.766117in}{2.638848in}}%
\pgfpathlineto{\pgfqpoint{4.772545in}{2.618480in}}%
\pgfpathlineto{\pgfqpoint{4.778794in}{2.598057in}}%
\pgfpathlineto{\pgfqpoint{4.784865in}{2.577580in}}%
\pgfpathlineto{\pgfqpoint{4.790757in}{2.557051in}}%
\pgfpathlineto{\pgfqpoint{4.796469in}{2.536471in}}%
\pgfpathlineto{\pgfqpoint{4.802002in}{2.515842in}}%
\pgfpathlineto{\pgfqpoint{4.807354in}{2.495166in}}%
\pgfpathlineto{\pgfqpoint{4.812526in}{2.474444in}}%
\pgfpathlineto{\pgfqpoint{4.817517in}{2.453677in}}%
\pgfpathlineto{\pgfqpoint{4.822327in}{2.432868in}}%
\pgfpathlineto{\pgfqpoint{4.826954in}{2.412017in}}%
\pgfpathlineto{\pgfqpoint{4.831400in}{2.391127in}}%
\pgfpathlineto{\pgfqpoint{4.835663in}{2.370199in}}%
\pgfpathlineto{\pgfqpoint{4.839743in}{2.349234in}}%
\pgfpathlineto{\pgfqpoint{4.843640in}{2.328235in}}%
\pgfpathlineto{\pgfqpoint{4.847354in}{2.307203in}}%
\pgfpathlineto{\pgfqpoint{4.850884in}{2.286138in}}%
\pgfpathlineto{\pgfqpoint{4.854230in}{2.265044in}}%
\pgfpathlineto{\pgfqpoint{4.929620in}{1.775387in}}%
\pgfpathlineto{\pgfqpoint{4.932598in}{1.754237in}}%
\pgfpathlineto{\pgfqpoint{4.935390in}{1.733063in}}%
\pgfpathlineto{\pgfqpoint{4.937998in}{1.711865in}}%
\pgfpathlineto{\pgfqpoint{4.940421in}{1.690645in}}%
\pgfpathlineto{\pgfqpoint{4.942659in}{1.669404in}}%
\pgfpathlineto{\pgfqpoint{4.944711in}{1.648145in}}%
\pgfpathlineto{\pgfqpoint{4.946577in}{1.626869in}}%
\pgfpathlineto{\pgfqpoint{4.948258in}{1.605578in}}%
\pgfpathlineto{\pgfqpoint{4.949753in}{1.584272in}}%
\pgfpathlineto{\pgfqpoint{4.951062in}{1.562954in}}%
\pgfpathlineto{\pgfqpoint{4.952185in}{1.541626in}}%
\pgfpathlineto{\pgfqpoint{4.953121in}{1.520289in}}%
\pgfpathlineto{\pgfqpoint{4.953872in}{1.498944in}}%
\pgfpathlineto{\pgfqpoint{4.954436in}{1.477594in}}%
\pgfpathlineto{\pgfqpoint{4.954814in}{1.456239in}}%
\pgfpathlineto{\pgfqpoint{4.955005in}{1.434882in}}%
\pgfpathlineto{\pgfqpoint{4.955010in}{1.413524in}}%
\pgfpathlineto{\pgfqpoint{4.954829in}{1.392167in}}%
\pgfpathlineto{\pgfqpoint{4.954461in}{1.370812in}}%
\pgfpathlineto{\pgfqpoint{4.953907in}{1.349462in}}%
\pgfpathlineto{\pgfqpoint{4.953167in}{1.328117in}}%
\pgfpathlineto{\pgfqpoint{4.952240in}{1.306779in}}%
\pgfpathlineto{\pgfqpoint{4.951128in}{1.285450in}}%
\pgfpathlineto{\pgfqpoint{4.949829in}{1.264132in}}%
\pgfpathlineto{\pgfqpoint{4.948344in}{1.242825in}}%
\pgfpathlineto{\pgfqpoint{4.946673in}{1.221533in}}%
\pgfpathlineto{\pgfqpoint{4.944817in}{1.200256in}}%
\pgfpathlineto{\pgfqpoint{4.942775in}{1.178996in}}%
\pgfpathlineto{\pgfqpoint{4.940548in}{1.157755in}}%
\pgfpathlineto{\pgfqpoint{4.938135in}{1.136533in}}%
\pgfpathlineto{\pgfqpoint{4.935537in}{1.115334in}}%
\pgfpathlineto{\pgfqpoint{4.932754in}{1.094158in}}%
\pgfpathlineto{\pgfqpoint{4.929787in}{1.073008in}}%
\pgfpathlineto{\pgfqpoint{4.926635in}{1.051884in}}%
\pgfpathlineto{\pgfqpoint{4.923299in}{1.030788in}}%
\pgfpathlineto{\pgfqpoint{4.919779in}{1.009722in}}%
\pgfpathlineto{\pgfqpoint{4.916075in}{0.988688in}}%
\pgfpathlineto{\pgfqpoint{4.912188in}{0.967687in}}%
\pgfpathlineto{\pgfqpoint{4.908117in}{0.946720in}}%
\pgfpathlineto{\pgfqpoint{4.460686in}{1.285802in}}%
\pgfpathlineto{\pgfqpoint{3.975508in}{1.463692in}}%
\pgfpathlineto{\pgfqpoint{3.481955in}{1.518535in}}%
\pgfpathlineto{\pgfqpoint{2.992359in}{1.491438in}}%
\pgfpathlineto{\pgfqpoint{2.507591in}{1.423623in}}%
\pgfpathlineto{\pgfqpoint{1.453387in}{1.261109in}}%
\pgfpathlineto{\pgfqpoint{1.381159in}{1.729644in}}%
\pgfpathclose%
\pgfusepath{stroke,fill}%
\end{pgfscope}%
\begin{pgfscope}%
\pgfpathrectangle{\pgfqpoint{1.077500in}{0.819137in}}{\pgfqpoint{6.680500in}{2.806826in}}%
\pgfusepath{clip}%
\pgfsetbuttcap%
\pgfsetmiterjoin%
\definecolor{currentfill}{rgb}{0.274510,0.509804,0.705882}%
\pgfsetfillcolor{currentfill}%
\pgfsetlinewidth{1.003750pt}%
\definecolor{currentstroke}{rgb}{0.000000,0.000000,0.000000}%
\pgfsetstrokecolor{currentstroke}%
\pgfsetdash{}{0pt}%
\pgfpathmoveto{\pgfqpoint{3.820010in}{2.079781in}}%
\pgfpathlineto{\pgfqpoint{4.871492in}{2.259064in}}%
\pgfpathlineto{\pgfqpoint{4.951173in}{1.791738in}}%
\pgfpathlineto{\pgfqpoint{3.899691in}{1.612456in}}%
\pgfpathclose%
\pgfusepath{stroke,fill}%
\end{pgfscope}%
\begin{pgfscope}%
\pgfpathrectangle{\pgfqpoint{1.077500in}{0.819137in}}{\pgfqpoint{6.680500in}{2.806826in}}%
\pgfusepath{clip}%
\pgfsetbuttcap%
\pgfsetmiterjoin%
\definecolor{currentfill}{rgb}{0.501961,0.501961,0.000000}%
\pgfsetfillcolor{currentfill}%
\pgfsetlinewidth{1.003750pt}%
\definecolor{currentstroke}{rgb}{0.000000,0.000000,0.000000}%
\pgfsetstrokecolor{currentstroke}%
\pgfsetdash{}{0pt}%
\pgfpathmoveto{\pgfqpoint{1.381159in}{1.729644in}}%
\pgfpathlineto{\pgfqpoint{2.435363in}{1.892158in}}%
\pgfpathlineto{\pgfqpoint{2.507591in}{1.423623in}}%
\pgfpathlineto{\pgfqpoint{1.453387in}{1.261109in}}%
\pgfpathclose%
\pgfusepath{stroke,fill}%
\end{pgfscope}%
\begin{pgfscope}%
\pgfpathrectangle{\pgfqpoint{1.077500in}{0.819137in}}{\pgfqpoint{6.680500in}{2.806826in}}%
\pgfusepath{clip}%
\pgfsetbuttcap%
\pgfsetmiterjoin%
\definecolor{currentfill}{rgb}{0.698039,0.133333,0.133333}%
\pgfsetfillcolor{currentfill}%
\pgfsetlinewidth{1.003750pt}%
\definecolor{currentstroke}{rgb}{0.000000,0.000000,0.000000}%
\pgfsetstrokecolor{currentstroke}%
\pgfsetdash{}{0pt}%
\pgfpathmoveto{\pgfqpoint{4.927718in}{1.787739in}}%
\pgfpathlineto{\pgfqpoint{3.899691in}{1.612456in}}%
\pgfpathlineto{\pgfqpoint{3.820010in}{2.079781in}}%
\pgfpathlineto{\pgfqpoint{4.855569in}{2.256349in}}%
\pgfpathclose%
\pgfusepath{stroke,fill}%
\end{pgfscope}%
\end{pgfpicture}%
\makeatother%
\endgroup%